\documentclass[11pt]{article}

\usepackage[final]{acl}

\usepackage{times}
\usepackage{latexsym}

\usepackage[T1]{fontenc}

\usepackage[utf8]{inputenc}

\usepackage{microtype}

\usepackage{silence}
\usepackage{inconsolata}

\usepackage{graphicx}

\usepackage{amsfonts} 
\usepackage{amsmath} 
\usepackage{framed} 
\usepackage{booktabs} 
\usepackage{nicematrix} 
\usepackage[table]{xcolor} 
\usepackage{arydshln} 
\usepackage[flushleft]{threeparttable} 
\usepackage{multirow} 
\usepackage{pifont} 
\usepackage{tcolorbox} 
\usepackage{enumitem} 
\usepackage{bbm}
\definecolor{forestgreen}{HTML}{008000} 
\definecolor{orng}{rgb}{1.0, 0.31, 0.0} 

\definecolor{AA}{HTML}{7B3294}
\definecolor{BB}{HTML}{E69F00}
\definecolor{CC}{HTML}{4D908E}
\definecolor{DD}{HTML}{6E6E6E}

\title{CliQ-RRG: Clinical-Knowledge Guided Disease-aware Visual–Textual Alignment for QA-Style Radiology Report Generation}

\author{
\textbf{Chandravardhan Singh Raghaw}$^{1}$,
\textbf{Shahid Shafi Dar}$^{1}$,
\textbf{Arnav Nishith Sharma}$^{2}$,\\
\textbf{Shrey Modi}$^{3}$,
\textbf{Nagendra Kumar}$^{1,*}$\\
$^{1}$Indian Institute of Technology Indore, Indore, India; $^{2}$University of Texas at Dallas, USA;\\
$^{3}$University of Wisconsin - Madison, USA\\
\texttt{\{phd2201101016, phd2201201004, nagendra\}@iiti.ac.in},\\
\texttt{axs230011@utdallas.edu},\
\texttt{modi7@wisc.edu}\\
$^{*}$Corresponding author
}

\begin{document}
\maketitle
\begin{abstract}
Radiology reports are essential for diagnostic reasoning and patient care, yet their manual preparation is time-consuming and cognitively demanding. Automatic radiology report generation (RRG) offers a scalable alternative, but existing models often produce lengthy, unstructured narratives that overlook diagnostic cues and multi-view information. We present \textbf{\underline{Cli}}nical-Knowledge Guided Disease-aware Visual–Textual Alignment for \textbf{\underline{Q}}A-Style \textbf{\underline{R}}adiology \textbf{\underline{R}}eport \textbf{\underline{G}}eneration (CliQ-RRG), a unified two-stage framework for interpretable and clinically grounded reporting. In Stage~1, CliQ-RRG employs a Disease-aware Visual–Textual Alignment module that aligns image and text representations using predicted disease embeddings, reinforced by a Prior-Guided Attention Module (PrAM) to capture multi-view dependencies across current and prior scans. In Stage~2, GraphRAG-guided retrieval augments intermediate textual representations with graph-grounded clinical evidence, and a large language model restructures them into concise, interpretable question–answer (QA) pairs with diagnostic summaries. Experiments on two public chest X-ray benchmarks demonstrate that CliQ-RRG consistently outperforms prior methods across both lexical and clinical metrics, generating accurate and clinically coherent QA-style radiology reports. Code is available at \url{https://github.com/chandravardhan-singh-raghaw/CliQ-RRG}.
\end{abstract}

\section{Introduction}
Radiology report generation (RRG) is vital for chest X-ray (CXR) interpretation, requiring substantial clinical expertise and reasoning~\cite{Jin2024promptmrg}. Manual interpretation, however, remains time-consuming and cognitively demanding, even for experts~\cite{Liu2025enhancercl, Park2025dart}. As imaging volumes rise, radiologists face increasing workload, often affecting report quality and diagnostic accuracy. Consequently, automated RRG has emerged as a promising solution~\cite{Hou2023organ, Liang2024divide, Luo2024textual}, yet most methods emphasize disease-specific regions, overlooking broader contextual information.
\begin{figure}[!b]
    \centering
    \includegraphics[width=0.9\columnwidth]{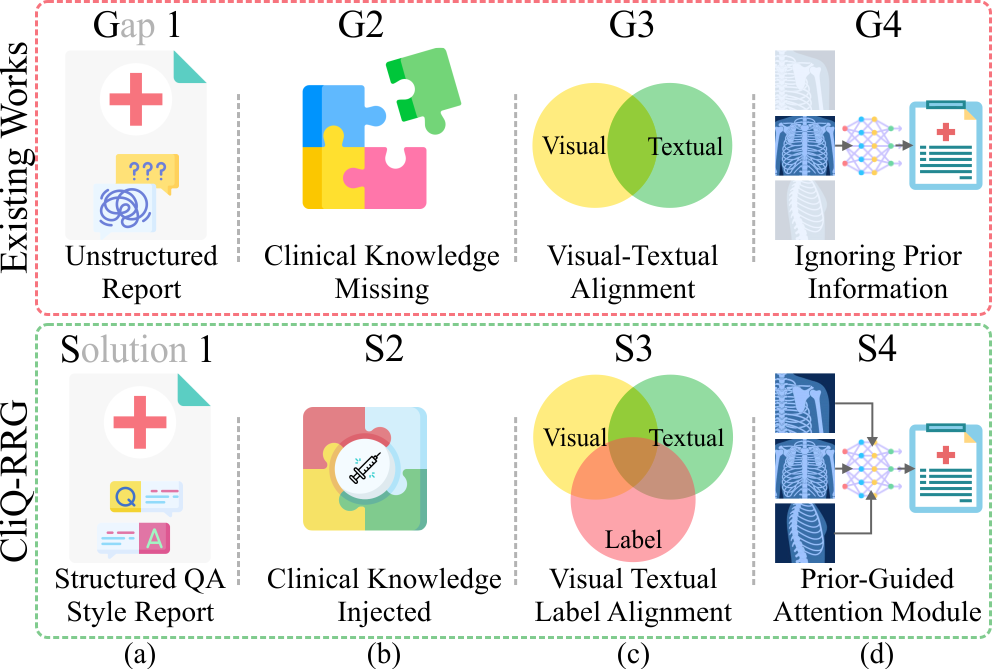}
    \caption{
    Comparison with existing methods. Existing approaches generate unstructured reports, overlook prior scans, lack clinical knowledge, and exhibit limited alignment. In contrast, CliQ-RRG generates QA-style reports with clinical knowledge enrichment, disease-aware alignment, and prior-guided attention.
    }
    \label{fig:intro}
\end{figure}

Recent studies have proposed advanced techniques to improve automatic RRG. \citet{Li2024kargen} integrates graph-enhanced and regional features to describe normal and abnormal findings, yet outputs remain lengthy and less interpretable. To embed clinical knowledge, \citet{Hou2025radar} and \citet{Sun2025fact} retrieve domain-specific information, but the generated texts lack diagnostic coherence. \citet{Huangetal2025cmeaa} aligns visual and textual features via a cross-modal adapter; however, the approach is inefficient and overlooks supervised labels that strengthen the alignment. Dual-view RRG methods~\cite{Chen2021cross, Yang2023radiology} differentiate radiographic views but miss fine anatomical and contextual relations across scans, limiting multi-view consistency in generated reports.

Despite progress in RRG~\cite{zhang2020radiology, yan2022clinical}, existing methods face four key limitations (Fig.~\ref{fig:intro}). (\textbf{G1}) Prior works~\cite{Jin2024promptmrg, Luo2024textual, Wang2025llmrg4} generate long free-text narratives, in which key findings are embedded within paragraphs rather than presented in a structured form. Existing studies have limitations of lengthy and inconsistent structured clinical text~\cite{kell2024question, pattnayak2025clinical, schwartz2011improving}. We therefore explore a QA-style reformulation to make findings explicit and easier to retrieve, aiming to improve structural clarity rather than replace narrative reports. (\textbf{G2}) Many studies~\cite{Yan2023see, Bu2024iekagen, Gu2024complex} rely on image–report pairs, which may yield fluent but underspecified descriptions, especially for subtle findings. We incorporate external clinical knowledge at the intermediate stage to promote clinically grounded expressions and strengthen alignment between visual evidence and generated text. (\textbf{G3}) Recent studies~\cite{Li2023dynamic, Shen2024man, Xiao2025mpo} align visual and textual modalities but neglect diagnostic labels as supervisory signals for joint embedding learning, unlike prompt-based approaches~\cite{Jin2024promptmrg}. We integrate predicted diagnostic labels directly into a contrastive alignment objective, structuring the joint visual–textual space based on diagnostic similarity. (\textbf{G4}) While several methods~\cite{Hou2023organ, Gu2025orid, Liu2025hcllm} leverage historical information, most standard approaches do not explicitly fuse prior scans at the visual representation level. We introduce prior-image fusion before alignment to capture temporal context within the learned feature space.

We present Clinical-Knowledge Guided Disease-aware Visual–Textual Alignment for QA-Style Radiology Report Generation (CliQ-RRG), a unified framework centered on two core contributions: disease-grounded GraphRAG retrieval and fine-tuned LLM-based QA restructuring. As shown in Fig.~\ref{fig:intro}, CliQ-RRG addresses limitations of prior RRG methods. (\textbf{S1}) It reformulates free-text reports into clinically interpretable QA-style outputs. (\textbf{S2}) It injects external clinical knowledge to enable reasoning beyond data-driven correlations. (\textbf{S3}) It aligns visual and textual representations via predicted disease embeddings to achieve semantically grounded features. (\textbf{S4}) It incorporates multi-view information from prior examinations through a prior-guided attention mechanism to enhance contextual understanding. Extensive experiments on \textbf{MIMIC-CXR} and \textbf{IU X-Ray} show that CliQ-RRG consistently outperforms state-of-the-art methods in radiology report generation.

\noindent Our main contributions are threefold:
\begin{itemize}
    \item To the best of our knowledge, CliQ-RRG is the first framework to leverage contrastive alignment for QA-style radiology report generation, restructuring unstructured narratives into concise, structured question-answer pairs.
    \item We introduce a Disease-aware Visual-Textual Alignment module that aligns image and text representations via predicted disease embeddings, guided by a Prior-Guided Attention Module for multi-view contextual integration.
    \item We design a GraphRAG-guided generation pipeline that enriches intermediate reports with graph-grounded clinical evidence and employs a large language model to synthesize the final QA-style report.
\end{itemize}

\section{Proposed Approach}
\subsection{Problem Formulation and Overview}
\begin{figure*}[!ht]
\centering
    \includegraphics[width=0.96\textwidth]{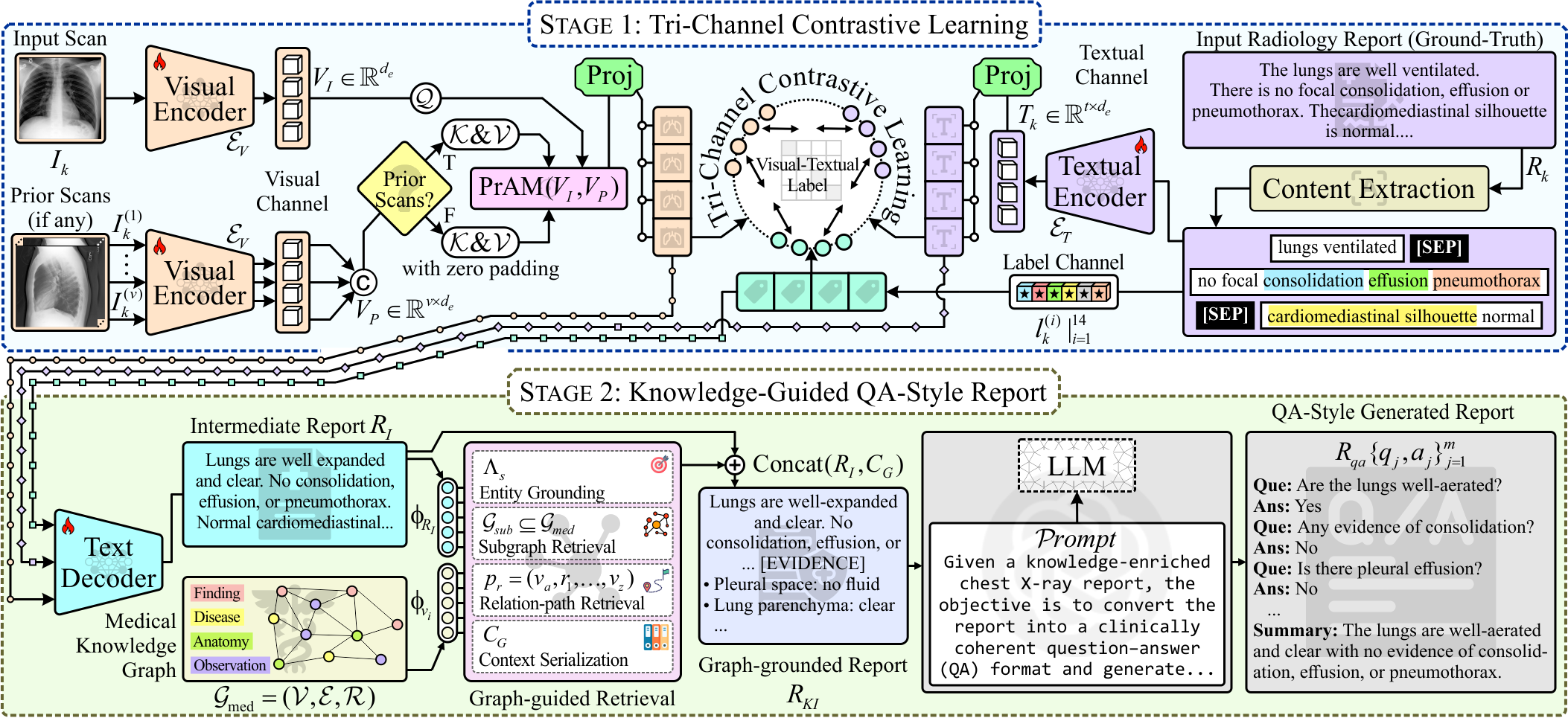} 
    \caption{
    Overview of the proposed CliQ-RRG framework. \textsc{Stage~1}: Disease-aware Visual--Textual Alignment aligns image and text representations using predicted disease embeddings, while the Prior-Guided Attention Module (PrAM) integrates multi-view information from current and prior scans. \textsc{Stage~2}: Knowledge-Guided QA-Style Report Generation enriches the intermediate report with graph-connected clinical evidence retrieved from a medical knowledge graph and employs a large language model to produce concise QA pairs and a diagnostic summary.}
\label{fig:cliqrrg}
\end{figure*}

Let $\mathcal{T} = \{(I_k, R_k, L_k)\}_{k=1}^n$ be the training set of $n$ studies, each with a CXR $I_k \in \mathbb{R}^{h \times w \times (v+1)}$, associated report $R_k$, and predicted disease labels $L_k$, where $v$ is the number of prior scans ($v=0$ if unavailable), and $h, w$ are image dimensions. Our objective is to generate a QA-style report $R_{qa} (q_i, a_i)$, where each question $q_i$ targets a clinical finding, and the answer $a_i \in \{\textrm{Yes, No}\}$ is grounded in visual evidence from $I$. The value $m$ is the total number of generated question–answer pairs. We hypothesize that aligning visual and textual representations with predicted disease labels allows our framework to generate QA-style radiology reports. We formalize the generation process as:
\setlength{\abovedisplayskip}{1pt}
\setlength{\belowdisplayskip}{1pt}
\setlength{\abovedisplayshortskip}{1pt}
\setlength{\belowdisplayshortskip}{1pt}
\begin{equation}
\label{eq:objective}
    R_{qa} (q_k, a_k) = \textrm{CliQ-RRG}(I, R, L)
\end{equation}
Our proposed framework, CliQ-RRG, operates in two stages, illustrated in Fig.~\ref{fig:cliqrrg}. \textsc{Stage} 1 employs a disease-aware visual–textual alignment module to align visual features from current and prior chest X-rays with textual representations and predicted disease labels in a unified embedding space. A prior-guided attention module further captures multi-view information from prior scans, enhancing temporal and anatomical consistency. In \textsc{Stage} 2, a text decoder generates an intermediate report, which we enhance through GraphRAG-guided retrieval of graph-grounded clinical evidence. Finally, an LLM restructures the knowledge-injected report into a concise QA-style format with question–answer pairs and a clinical summary.

\subsection{\textsc{Stage} 1: Disease-aware Visual–Textual Alignment}
\noindent\textbf{Visual Channel:}  
Let $I_k$ denote the current CXR and ${I_k^{(j)}}_{j=1}^{v}$ the set of $v$ prior scans for the $k^{\text{th}}$ patient ($v \geq 0$). We use the BioMedCLIP~\cite{Zhang2025biomedclip} visual encoder $\mathcal{E}_V(\cdot)$ to extract image features. The current view is encoded as $V_I = \mathcal{E}_V(I_k)$, and each prior view as $V_P^{(j)} = \mathcal{E}_V(I_k^{(j)})$. Prior features are concatenated to form a contextual representation $V_P = \mathrm{Concat}(V_P^{(1)}, \ldots, V_P^{(v)})$, zero-padded when $v = 0$. The visual channel representation is defined as $V_{ch} = \mathrm{PrAM}(V_I, V_P)$, where PrAM fuses the current image features $V_I$ and prior image features $V_P$, capturing both current and multi-view information.

\noindent\textbf{Prior-guided Attention Module:} 
To handle variable image counts across studies, which complicate concatenation and hinder cross-modal alignment, we introduce the Prior-Guided Attention Module (PrAM). PrAM integrates multi-view information from prior scans through a cross-attention mechanism~\cite{Vaswani2017attention}. The current image feature $V_I \in \mathbb{R}^{d_e}$ serves as the query $\mathcal{Q}$, and the prior features $V_P \in \mathbb{R}^{v \times d_e}$ act as both keys $\mathcal{K}$ and values $\mathcal{V}$, where $d_e$ is the shared embedding dimension. The output is computed as:
\setlength{\abovedisplayskip}{1pt}
\setlength{\belowdisplayskip}{1pt}
\setlength{\abovedisplayshortskip}{1pt}
\setlength{\belowdisplayshortskip}{1pt}
\begin{equation}
\label{eq:pram}
    \mathrm{PrAM}(V_I, V_P) = \mathrm{softmax}\!\left(\frac{V_I V_P^\top}{\sqrt{d_e}}\right) \cdot V_P
\end{equation}
producing a unified representation that preserves multi-view context while maintaining consistent dimensionality for contrastive alignment. PrAM does not explicitly encode the temporal order or time intervals between prior scans; hence, it captures historical context but does not explicitly model longitudinal progression patterns.

\noindent\textbf{Textual Channel:} 
To process radiology reports $R_k$, we construct a textual channel $T_{ch}$ that encodes clinical information. Following \cite{Yan2023see, Liu2025enhancercl, liu2024factual}, we extract a sequence of $t_c$ salient clinical phrases from $R_k$, separated by \texttt{[SEP]} tokens, as shown in Fig.~\ref{fig:cliqrrg}. This sequence is passed through a textual encoder $\mathcal{E}_T(\cdot)$, instantiated as CXR-BERT \cite{Boecking2022cxrbert}, yielding textual features $T_k = \mathcal{E}_T(R_k) \in \mathbb{R}^{{t_c} \times d_e}$. These representations enable alignment with visual and label modalities during contrastive learning. During inference, ground-truth reports are unavailable; therefore, $T_{ch}$ is not used. The decoder relies solely on the visual channel $V_{ch}$ and predicted diagnostic labels $L_{ch}$. Since $V_{ch}$ has been aligned with $T_{ch}$ during training, it encodes the necessary semantic information to guide generation.

\noindent\textbf{Disease Classifier:} 
To capture disease-relevant supervision, we design a disease classifier that predicts diagnostic labels ($L_{ch}$) from the visual channel features $V_{ch} \in \mathbb{R}^{d_e}$. Rather than a single softmax jointly over all 13 disease categories, the classifier computes, for each disease category $i$, a probability distribution over CheXbert's four states (uncertain, negative, positive, not mentioned) independently, using a cross-attention mechanism~\cite{Vaswani2017attention}:
\setlength{\abovedisplayskip}{1pt}
\setlength{\belowdisplayskip}{1pt}
\setlength{\abovedisplayshortskip}{1pt}
\setlength{\belowdisplayshortskip}{1pt}
\begin{gather}
\hat{Y}^{(i)} = \mathrm{Softmax}\left(\frac{V_{ch} \Phi^{(i)\top}}{\sqrt{d_e}}\right), \hat{Y}^{(i)} \in \mathbb{R}^{4} \\
L_{ch}^{(i)} = \arg\max_{s \in \{1,2,3,4\}} \hat{Y}^{(i,s)} \\
\mathcal{L}_{CE} = \sum_{i=1}^{13} \mathrm{CrossEntropy}(\hat{Y}^{(i)}, L_{gt}^{(i)})
\end{gather}
where $\Phi \in \mathbb{R}^{(d \times 4) \times d_e}$ is a learnable disease embedding matrix and $L_{gt}$ represents the ground-truth disease labels extracted using CheXbert~\cite{Smit2020chexbert}. Each label $l_k^{(i)} \in \{-1, 0, 1, 2\}$ denotes uncertainty, negative, positive, or not mentioned, respectively. This yields genuinely multi-label, multi-state prediction, with each of the 13 disease categories classified independently over its own 4-state distribution rather than one distribution shared across all diseases. We retain the first 13 disease categories to identify reports with positive or uncertain findings. Appendix~\ref{sec:perstate} reports per-state classification performance validating this formulation.

\begin{figure}[!htbp]
    \centering
    \includegraphics[width=0.9\columnwidth]{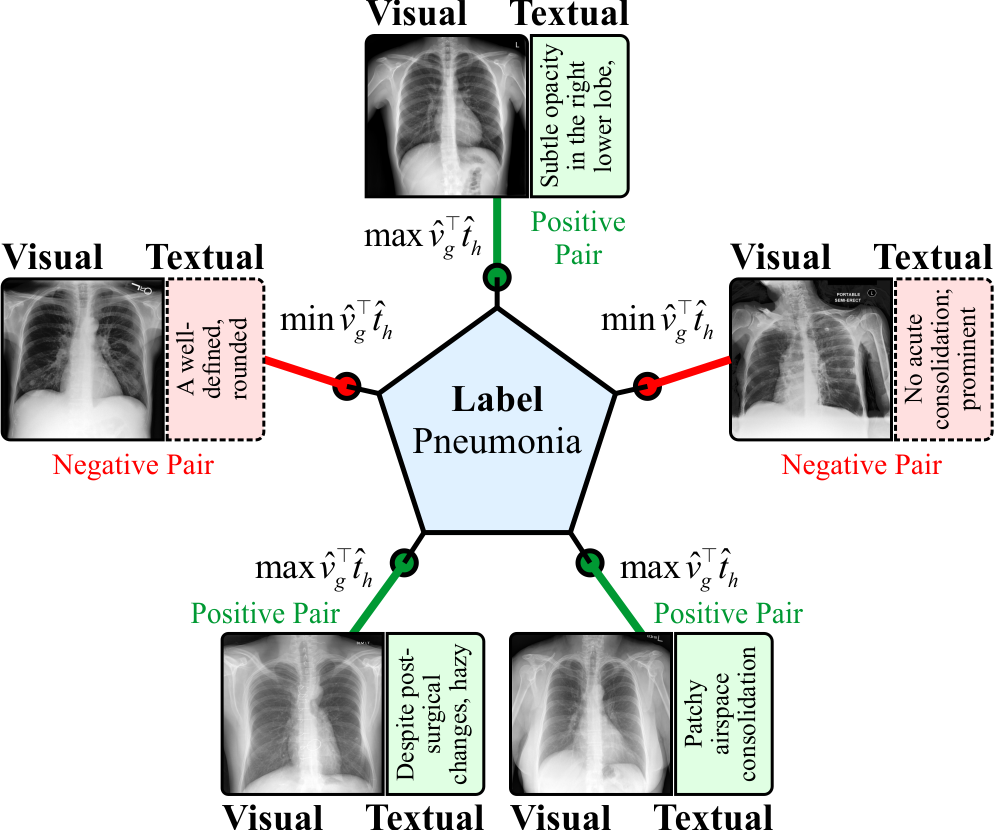}
    \caption{
    Disease-aware Visual--Textual Contrast (Di-VTC). Positive visual--textual pairs with shared disease labels (green) are pulled closer, while unrelated pairs (red) are pushed apart, enforcing disease-consistent alignment in the shared embedding space.
    }
    \label{fig:divtc}
\end{figure}

\noindent\textbf{Disease-aware Visual–Textual Contrast:} 
Building on prior contrastive and alignment-based RRG methods that align visual and textual representations through learned knowledge bases or region-level correspondence~\cite{Yang2023radiology, Chen2024adamatchcyclic}, we propose Disease-aware Visual–Textual Contrast (Di-VTC), which adapts contrastive learning to the generative setting of QA-style radiology report generation. While prior work applies contrastive learning primarily in discriminative settings, Di-VTC learns the parameters $\theta_V$ and $\theta_T$ of the visual ($\mathcal{E}_V$) and textual ($\mathcal{E}_T$) encoders to align visual ($V_{ch}$) and textual ($T_{ch}$) representations in a shared embedding space. For samples $g$ and $h$ in a batch $B$, the channel features are projected and $L_2$-normalized to yield $\hat{v}_g {=} \text{Norm}(\text{Proj}(V_{ch}^{(g)}))$ and $\hat{t}_h {=} \text{Norm}(\text{Proj}(T_{ch}^{(h)}))$. A pair is treated as positive if their predicted diagnostic labels are similar, denoted as $l_g {\approx} l_h$, where two samples are considered similar when their disease label vectors are identical or satisfy a predefined similarity threshold; otherwise, the pair is treated as negative. As shown in Fig.~\ref{fig:divtc}, Di-VTC employs a push–pull dynamic that maximizes $\hat{v}_g^{\top} \hat{t}_h$ for positive pairs and minimizes it for negative pairs. The model is optimized using a bidirectional contrastive loss $\mathcal{L}_{\text{Di-VTC}}$ over $\theta_V$ and $\theta_T$.
\setlength{\abovedisplayskip}{1pt}
\setlength{\belowdisplayskip}{1pt}
\setlength{\abovedisplayshortskip}{1pt}
\setlength{\belowdisplayshortskip}{1pt}
\begin{equation}
\label{eq:divtc}
    \min_{\{\theta_V, \theta_T\}} \mathcal{L}_{\text{Di-VTC}} = \mathcal{L}_{v \mapsto t} + \mathcal{L}_{t \mapsto v}
\end{equation}
The $\mathcal{L}_{\text{Di-VTC}}$ loss combines a visual-to-textual loss $(\mathcal{L}_{v \mapsto t})$ and a textual-to-visual loss $(\mathcal{L}_{t \mapsto v})$:
\begin{gather}
\label{eq:v2t_loss}
    \mathcal{L}_{v \mapsto t}\! = \!
    - \sum_{g \in B} \!\frac{1}{|\mathrm{p}(g)|} \!
    \sum_{i \in \mathrm{p}(g)} \!
    \log \!
    \frac{\exp(\hat{v}_{g}^{\top} \hat{t}_{i} / \tau)}
    {\sum_{h \in B} \exp(\hat{v}_{g}^{\top} \hat{t}_{h} / \tau)}
\end{gather}
\begin{gather}
\label{eq:t2v_loss}
    \mathcal{L}_{t \mapsto v}\! = \!
    - \sum_{h \in B}\! \frac{1}{|\mathrm{p}(h)|} \!
    \sum_{i \in \mathrm{p}(h)} \!
    \log \!
    \frac{\exp(\hat{v}_{i}^{\top} \hat{t}_{h} / \tau)}
    {\sum_{g \in B} \exp(\hat{v}_{g}^{\top} \hat{t}_{h} / \tau)}
\end{gather}
where $\mathrm{p}(\cdot)$ denotes the set of positive pair indices for a given anchor sample, and $\tau$ is a temperature hyperparameter. The Di-VTC framework aligns visual and textual representations using similarity in predicted disease embeddings as supervision, enabling clinically grounded and coherent QA-style report generation.

\subsection{\textsc{Stage} 2: Knowledge-Guided QA-Style Report}

\noindent\textbf{Intermediate Report Generation:}
We generate the intermediate report $R_I$ using a text decoder conditioned on visual channel $V_{ch}$, textual channel $T_{ch}$, and predicted diagnostic labels $L_{ch}$ from \textsc{Stage} 1. For each $k^{\text{th}}$ study, the decoder predicts token $R_k^t$ autoregressively using preceding tokens $R_k^{<t}$ and multi-channel representations. The decoder is optimized using negative log-likelihood loss:
\setlength{\abovedisplayskip}{1pt}
\setlength{\belowdisplayskip}{1pt}
\setlength{\abovedisplayshortskip}{1pt}
\setlength{\belowdisplayshortskip}{1pt}
\begin{equation}
\label{eq:ir_loss}
    \mathcal{L}_G = -\sum_{t=1}^{l}\log (R_k^t|R_k^{<t},V_{ch},T_{ch},L_{ch})
\end{equation}
where $l$ denotes report length and $p(R_k^t|\cdot)$ denotes the probability of generating token $R_k^t$ conditioned on preceding tokens and decoder inputs. The loss $\mathcal{L}_G$ encourages clinically coherent report generation aligned with $(V_{ch},T_{ch},L_{ch})$.

\smallskip
\noindent\textbf{Graph-guided Retrieval:}
Unlike prior knowledge-injected RRG frameworks that incorporate external clinical text during decoding~\cite{Huang2023kiut,Bu2024iekagen}, CliQ-RRG performs Graph Retrieval-Augmented Generation (GraphRAG) before QA synthesis. Instead of retrieving isolated knowledge fragments, our framework retrieves clinically connected evidence through entity grounding, subgraph retrieval, and relation-path reasoning.

We organize the external clinical repository (Appendix~\ref{sec:knowledgebase}) as a medical knowledge graph:
\setlength{\abovedisplayskip}{1pt}
\setlength{\belowdisplayskip}{1pt}
\setlength{\abovedisplayshortskip}{1pt}
\setlength{\belowdisplayshortskip}{1pt}
\begin{equation}
\mathcal{G}_{med}
=
(\mathcal{V},\mathcal{E},\mathcal{R})
\end{equation}
where $\mathcal{V}$ denotes clinical entities including anatomy, findings, and diseases. Edges $e=(v_a,r,v_b)\in\mathcal{E}$ encode medical relations $r\in\mathcal{R}$.

Given intermediate report $R_I$, which serves as the retrieval query, and predicted disease labels $L_{ch}$ from Stage 1, we perform entity grounding by projecting report and graph entities into a shared semantic space. Intermediate reports are encoded using CXR-BERT to obtain query representation $\phi_{R_I}$, while graph entities are initialized using BioWordVec embeddings $\phi_{v_i}$. Graph retrieval relevance score $S(v_i)$ for graph entity $v_i$ is computed as:
\setlength{\abovedisplayskip}{1pt}
\setlength{\belowdisplayskip}{1pt}
\setlength{\abovedisplayshortskip}{1pt}
\setlength{\belowdisplayshortskip}{1pt}
\begin{equation}
\label{eq:graphrank}
\begin{aligned}
    S(v_i)  &= \alpha\,\mathrm{sim} (\phi_{R_I},\phi_{v_i})\\
            &\quad+(1-\alpha)\,\mathbb{I}\big(L_{ch}[\mathcal{M}(v_i)]\ge\tau\big)
\end{aligned}
\end{equation}
where $\mathrm{sim}(\cdot)$ denotes cosine similarity, $\mathbb{I}(\cdot)$ is an indicator function returning $1$ if the mapped label exceeds $\tau$ and $0$ otherwise, $\alpha$ balances semantic and diagnostic signals, and $\mathcal{M}(\cdot)$ maps graph entities to disease categories.

Top-ranked entities initialize retrieval seeds $\Lambda_s$. Starting from $\Lambda_s$, bounded breadth-first traversal retrieves evidence subgraph $\mathcal{G}_{sub}\subseteq\mathcal{G}_{med}$ to preserve clinically relevant dependencies while suppressing irrelevant evidence propagation. GraphRAG further retrieves multi-hop clinical relation paths
\setlength{\abovedisplayskip}{1pt}
\setlength{\belowdisplayskip}{1pt}
\setlength{\abovedisplayshortskip}{1pt}
\setlength{\belowdisplayshortskip}{1pt}
\begin{equation}
    p_r = (v_a,r_1,v_b,\dots,r_k,v_z)
\end{equation}
where $(v_m,r_k,v_n)\in\mathcal{E}$ denotes valid graph connectivity.
Retrieved paths are ranked and serialized into graph evidence context $C_G$. Graph-enhanced context is then obtained as:
\setlength{\abovedisplayskip}{1pt}
\setlength{\belowdisplayskip}{1pt}
\setlength{\abovedisplayshortskip}{1pt}
\setlength{\belowdisplayshortskip}{1pt}
\begin{equation}
    R_{KI} = \mathrm{Concat}(R_I,C_G)
\end{equation}
where $R_{KI}$ denotes graph-enhanced context for QA-style generation. Implementation details are provided in Appendix~\ref{sec:graphretrieval}.

\smallskip
\noindent\textbf{QA-Style Report Generation:}
We employ fine-tuned Llama-3-8B-Instruct, denoted by $\mathcal{F}_{qa}(\cdot)$, to restructure graph-enhanced report $R_{KI}^{(k)}$ into clinically meaningful QA pairs $R_{qa}^{(k)}$ and a concise summary, where $k$ indexes the study. A predefined $\mathcal{P}rompt$ guides $\mathcal{F}_{qa}(\cdot)$ to generate target format:
\begin{tcolorbox}[
    colback=black!7, 
    halign title=flush center, 
    colframe=black!80, 
    title=\footnotesize{\textsc{Prompt}},
    top=2pt, bottom=2pt, left=2pt, right=2pt, 
    boxsep=2pt
]
\scriptsize
\centering
\texttt{Given a knowledge-enriched chest X-ray report, the objective is to convert the report into a clinically coherent question–answer (QA) format and generate a concise diagnostic summary. This involves decomposing the report into short, interpretable QA pairs, followed by synthesizing a summary of the key findings.}
\end{tcolorbox}
\noindent We formalize QA generation as:
\begin{equation}
\label{eq:qareport}
    R_{qa}^{(k)} = \mathcal{F}_{qa}\left(R_{KI}^{(k)},\mathcal{P}rompt\right)
\end{equation}
The output $R_{qa}^{(k)}=\{(q_j,a_j)\}_{j=1}^{m}$ contains clinically interpretable QA pairs, where $m$ denotes the number of QA pairs, $q_j$ denotes a clinical question, and $a_j\in\{\mathrm{Yes,No}\}$ denotes its answer, followed by a concise diagnostic summary.

\subsection{Learning Objective}
We train the framework by jointly optimizing the disease-aware visual-textual contrastive, disease classification, and report generation objectives. The total loss is defined as:
\setlength{\abovedisplayskip}{2pt}
\setlength{\belowdisplayskip}{2pt}
\setlength{\abovedisplayshortskip}{2pt}
\setlength{\belowdisplayshortskip}{2pt}
\begin{equation}
\label{eq:total_loss}
    \mathcal{L}_{\mathrm{total}} 
    = \mathcal{L}_{\mathrm{Di\text{-}VTC}} 
    + \lambda_{1} \cdot \mathcal{L}_{\mathrm{G}} 
    + \lambda_{2} \cdot \mathcal{L}_{\mathrm{CE}}
\end{equation}
where $\mathcal{L}_{\mathrm{Di\text{-}VTC}}$ denotes the Disease-aware Visual--Textual Contrast loss, $\mathcal{L}_{\mathrm{CE}}$ is the disease classification loss computed using cross-entropy between predicted and CheXbert-derived labels, and $\mathcal{L}_{\mathrm{G}}$ represents the intermediate report generation loss. The coefficients $\lambda_{1}$ and $\lambda_{2}$ balance the contributions of the generative and classification objectives, and are empirically set to $1.0$ for equal weighting.

\begin{table*}[ht]
    \scriptsize
    \setlength{\tabcolsep}{2pt}
    \centering
    \caption{Quantitative comparison of CliQ-RRG with SOTA methods on MIMIC-CXR and IU X-Ray. $\dagger$: Results from published papers; $\star$: Results reproduced from released code. \textbf{Bold}: best and \underline{underlined}: second-best scores.}        
    \begin{threeparttable}  
    \begin{NiceTabular*}{\textwidth}{@{\extracolsep{\fill}} l l l cccccc cccccc @{}}
        \toprule[1.25pt]
        \textbf{Type} & \textbf{Method} & \textbf{Published}
        & \multicolumn{6}{c}{\textbf{MIMIC-CXR}} 
        & \multicolumn{6}{c}{\textbf{IU X-Ray}} \\
        \cmidrule(lr){4-9} \cmidrule(lr){10-15}
         
        & & 
        & \textbf{B1} & \textbf{B4} & \textbf{MTR} & \textbf{RG-L} & \textbf{C} & \textbf{BS$^\#$}
        & \textbf{B1} & \textbf{B4} & \textbf{MTR} & \textbf{RG-L} & \textbf{C} & \textbf{BS$^\#$} \\        
        \midrule        
        \midrule
        I & R2GenRL$^\star$ \cite{Qin2022r2genrl} & ACL'22 
        & 0.392 & 0.113 & 0.149 & 0.275 & 0.298 & 0.852 
        & 0.487 & 0.177 & 0.205 & 0.377 & 0.452 & 0.835 \\
        
        & DCL$^\star$ \cite{Li2023dynamic} & CVPR'23 
        & 0.384 & 0.114 & 0.147 & 0.278 & 0.301 & 0.844 
        & 0.483 & 0.171 & 0.201 & 0.391 & 0.461 & 0.838 \\
        
        & MAN$^\dagger$ \cite{Shen2024man} & AAAI'24 
        & 0.396 & 0.115 & 0.151 & 0.274 & -- & -- 
        & 0.501 & 0.170 & 0.213 & 0.386 & -- & -- \\

        & CoFE$^\star$ \cite{Li2024cofe} & ECCV'24 
        & 0.393 & 0.133 & 0.172 & 0.316 & \underline{0.385} & 0.860
        & 0.406 & 0.177 & 0.199 & 0.423 & 0.508 & 0.849 \\

        & SEI$^\star$ \cite{Liu2024sei} & MICCAI'24 
        & 0.381 & 0.119 & 0.148 & 0.306 & 0.328 & 0.841 
        & 0.486 & 0.173 & 0.209 & 0.395 & 0.472 & 0.837 \\

        & DART$^\dagger$ \cite{Park2025dart} & CVPR'25 
        & 0.437 & 0.137 & 0.175 & 0.310 & -- & -- 
        & 0.486 & 0.208 & 0.205 & 0.411 & -- & -- \\        

        \cdashline{1-15}[0.5pt/2pt]\noalign{\vskip 0.5ex}

        II & KiUT$^\dagger$ \cite{Huang2023kiut} & CVPR'23 
        & 0.393 & 0.113 & 0.160 & 0.285 & -- & -- 
        & 0.525 & 0.185 & \textbf{0.242} & 0.409 & -- & -- \\

        & EKAGen$^\dagger$ \cite{Bu2024iekagen} & CVPR'24 
        & 0.419 & 0.119 & 0.157 & 0.287 & -- & -- 
        & 0.526 & 0.203 & 0.214 & 0.404 & -- & -- \\
        
        & PromptMRG$^\dagger$ \cite{Jin2024promptmrg} & AAAI'24 
        & 0.398 & 0.112 & 0.157 & 0.268 & -- & -- 
        & 0.401 & 0.098 & 0.160 & 0.281 & -- & -- \\

        & RADAR$^\star$ \cite{Hou2025radar} & ACL'25 
        & \underline{0.493} & \underline{0.277} & 0.170 & 0.304 & -- & \underline{0.897}
        & 0.411 & 0.119 & 0.171 & 0.293 & -- & 0.840 \\                 
        & REVTAF$^\dagger$ \cite{Zhou2025revtaf} & ICCV'25 
        & 0.465 & 0.182 & \textbf{0.199} & \textbf{0.336} & -- & -- 
        & 0.420 & 0.107 & 0.176 & 0.309 & -- & -- \\

        \cdashline{1-15}[0.5pt/2pt]\noalign{\vskip 0.5ex}
        
        III & R2-LLM$^\star$ \cite{Liu2024r2llm} & AAAI'24 
        & 0.411 & 0.132 & 0.179 & 0.288 & 0.354 & 0.873
        & 0.487 & 0.178 & 0.215 & 0.401 & 0.485 & 0.846 \\
        
        & AdaMatch-Cyclic$^\dagger$ \cite{Chen2024adamatchcyclic} & ACL'24 
        & 0.379 & 0.101 & 0.163 & 0.286 & -- & -- 
        & 0.416 & 0.145 & 0.162 & 0.366 & -- & -- \\        
        
        & KARGEN$^\dagger$ \cite{Li2024kargen} & MICCAI'24 
        & 0.417 & 0.140 & 0.165 & 0.305 & 0.289 & -- 
        & 0.490 & 0.180 & 0.218 & 0.385 & 0.491 & -- \\        
        
        & LLM-RG4$^\star$ \cite{Wang2025llmrg4} & AAAI'25 
        & 0.384 & 0.136 & 0.155 & 0.322 & 0.372 & 0.849
        & 0.442 & 0.179 & 0.192 & 0.403 & 0.479 & \underline{0.848} \\

        & DAD-RG$^\dagger$ \cite{Wu26dadsrg} & AAAI'26 
        & 0.428 & 0.131 & -- & 0.291 & 0.232 & -- & 0.495 & 0.187 & -- & 0.384 & 0.634 & -- \\

        \cdashline{1-15}[0.5pt/2pt]\noalign{\vskip 0.5ex}
        
        IV & MPO$^\dagger$ \cite{Xiao2025mpo} & AAAI'25 
        & 0.416 & 0.139 & 0.162 & 0.309 & -- & -- 
        & \underline{0.548} & 0.209 & 0.224 & 0.415 & -- & -- \\ 

        & HC-LLM$^\star$ \cite{Liu2025hcllm} & AAAI'25 
        & 0.478 & 0.215 & 0.166 & 0.298 & 0.362 & 0.864 
        & 0.532 & 0.199 & 0.221 & 0.423 & 0.503 & 0.839 \\
        
        & ORID$^\dagger$ \cite{Gu2025orid} & WACV'25 
        & 0.386 & 0.117 & 0.150 & 0.284 & -- & -- 
        & 0.501 & 0.198 & 0.211 & 0.400 & -- & -- \\

        \cdashline{1-15}[0.5pt/2pt]\noalign{\vskip 0.5ex}
        
        V & CoD$^\dagger$ \cite{Jin2025codvqa} & TMI'25 
        & 0.412 & 0.129 & -- & 0.286 & -- & -- 
        & 0.403 & 0.091 & -- & 0.288 & -- & -- \\ 

        \cdashline{1-15}[0.5pt/2pt]\noalign{\vskip 0.5ex}
        
        VI & RA-RRG$^\dagger$ \cite{Park2026rarrg} & ACL'26 
        & 0.379 & 0.080 & -- & 0.249 & -- & -- & 0.363 & 0.067 & -- & 0.272 & -- & -- \\

        \cdashline{1-15}[0.5pt/2pt]\noalign{\vskip 0.5ex}
        
        VII & MAIRA-2$^\star$ \cite{bannur2024maira} & arXiv'24 
        & 0.401 & 0.128 & 0.168 & 0.296 & -- & 0.855 & -- & -- & -- & -- & -- & -- \\

        & MedGemma$^\star$ \cite{sellergren2025medgemma} & arXiv'25
        & 0.385 & 0.118 & 0.159 & 0.281 & -- & 0.847 & 0.412 & 0.145 & 0.187 & 0.362 & -- & 0.831 \\
                        
        \cdashline{1-15}[0.5pt/2pt]\noalign{\vskip 0.5ex}
        \rowcolor{black!7}
        Ours & CliQ-RRG (Graph-guided) &  
        & \textbf{0.502} & \textbf{0.283} & \underline{0.193} & \textbf{0.336} & \textbf{0.396} & \textbf{0.898}
        & \textbf{0.565} & \textbf{0.221} & \underline{0.225} & \textbf{0.438} & \underline{0.589} & \textbf{0.885} \\
        
        \bottomrule[1.25pt]
    \end{NiceTabular*}
        \begin{tablenotes}
    \scriptsize
    \centering
    \item I: Contrastive-based; II: Knowledge Injected-based; III: LLM-based, IV: Multi View-based; V: QA Style-based; VI: RAG-based; VII: MLLM-based; $^\#$: BERT scores are reported for $\star$ methods using reproduced outputs; All reproduced methods use the same test set for fair comparison
    \end{tablenotes}
    \end{threeparttable}
    \label{tab:sota}
\end{table*}

\begin{table}[!ht]
\scriptsize
\centering
\caption{
Comparison of CE metrics of CliQ-RRG on the MIMIC-CXR dataset. $\dagger$: Results from published papers. \textbf{Bold}: best and \underline{underlined}: second-best scores.
}
    \begin{NiceTabular*}{\columnwidth}{@{\extracolsep{\fill}} l@{} ccc}
        \toprule[1.25pt]
        \textbf{Method} & \textbf{Pre} & \textbf{Rec} & \textbf{F1} \\
        \midrule
        \midrule
        R2GenRL$^\dagger$ \cite{Qin2022r2genrl} & 0.342 & 0.294 & 0.292 \\        
        DCL$^\dagger$ \cite{Li2023dynamic} & 0.471 & 0.352 & 0.373 \\
        MAN$^\dagger$ \cite{Shen2024man} & 0.411 & 0.398 & 0.389 \\
        CoFE$^\dagger$ \cite{Li2024cofe} & 0.489 & 0.370 & 0.405 \\
        DART$^\dagger$ \cite{Park2025dart} & 0.533 & 0.520 & 0.546 \\

        \cdashline{1-4}[0.5pt/2pt]\noalign{\vskip 0.5ex}

        KiUT$^\dagger$ \cite{Huang2023kiut} & 0.371 & 0.318 & 0.321 \\
        EKAGen$^\dagger$ \cite{Bu2024iekagen} & 0.517 & 0.483 & 0.499 \\
        PromptMRG$^\dagger$ \cite{Jin2024promptmrg} & 0.501 & 0.509 & 0.476 \\
        REVTAF$^\dagger$ \cite{Zhou2025revtaf} & \textbf{0.628} & \underline{0.613} & \underline{0.592} \\

        \cdashline{1-4}[0.5pt/2pt]\noalign{\vskip 0.5ex}

        R2-LLM$^\dagger$ \cite{Liu2024r2llm} & 0.465 & 0.482 & 0.473 \\
        LLM-RG4$^\dagger$ \cite{Wang2025llmrg4} & 0.583 & 0.593 & 0.588 \\

        \cdashline{1-4}[0.5pt/2pt]\noalign{\vskip 0.5ex}
        
        MPO$^\dagger$ \cite{Xiao2025mpo} & 0.436 & 0.376 & 0.353 \\   
        ORID$^\dagger$ \cite{Gu2025orid} & 0.435 & 0.295 & 0.352 \\

        \cdashline{1-4}[0.5pt/2pt]\noalign{\vskip 0.5ex}        
        
        CoD$^\dagger$ \cite{Jin2025codvqa} & 0.487 & 0.521 & 0.479 \\
        
        \cdashline{1-4}[0.5pt/2pt]\noalign{\vskip 0.5ex}
        \rowcolor{black!7}
        CliQ-RRG (Graph-guided) & \underline{0.601} & \textbf{0.621} & \textbf{0.610} \\
        \cdashline{1-4}[0.5pt/2pt]\noalign{\vskip 0.5ex}
        \textit{Disease Classifier} &\textit{0.496} &\textit{0.515} &\textit{0.505} \\ 
        \bottomrule[1.25pt]
    \end{NiceTabular*}
    \label{tab:ce}
\end{table}

\section{Experimental Setting}
\noindent\textbf{Datasets:}
We evaluate CliQ-RRG on public datasets: (1) MIMIC-CXR \cite{johnson2019mimicjpg}, the largest RRG dataset with 377{,}110 scans and 227{,}827 reports, using official train/validation/test split \cite{chen2020rr, Park2025dart}; and (2) IU X-Ray \cite{DemnerFushman2015iu}, a smaller set with 7{,}470 frontal and lateral scans and 3{,}955 reports, divided into 7:2:1 splits following prior work \cite{chen2020rr, Park2025dart}.

\smallskip
\noindent\textbf{Metrics:} 
Following prior work~\cite{Jin2024promptmrg, Park2025dart}, narrative reports are evaluated using NLG metrics: BLEU (B)~\cite{papineni2001bleu}, METEOR (MTR)~\cite{Denkowski2011}, ROUGE-L (RG-L)~\cite{lin2004rouge}, CIDEr (C)~\cite{Vedantam2015cider}, and BERTScore (BS)~\cite{Zhang2019bertscore}. Clinical Efficacy (CE) is assessed using CheXbert~\cite{Smit2020chexbert} F1, Precision (Pre), and Recall (Rec). For QA pairs, we follow recent studies~\cite{Manes2024kqa, Kim2024medexqa} and evaluate semantic (BS) and clinical metrics (CheXbert-F1, RadGraph-F1~\cite{jain2021radgraph}). Inter-annotator agreement is measured using pairwise agreement and Fleiss’ kappa $(\kappa_F)$.

\smallskip
\noindent\textbf{Implementation Details}
The framework is implemented in TensorFlow and trained on an NVIDIA 64\,GB GPU using the AdamW optimizer~\cite{Loshchilov2019adam}. We train for $30$ epochs with batch size $16$, learning rate $3\mathrm{e}{-5}$, and weight decay $0.01$. Additional implementation details are provided in Appendix~\ref{sec:implementation}.

\begin{table*}[!ht]
    \scriptsize
    \setlength{\tabcolsep}{2pt}
    \centering
    \caption{
    Ablation analysis on MIMIC-CXR showing incremental performance gains from integrating PrAM, Di-VTC (across $V_{ch}$--$T_{ch}$ and $L_{ch}$), and $\mathcal{G}_{med}$ into the base model. Results are reported as mean $\pm$ std over five runs; * indicates statistically significant improvement over the base model ($p\le0.05$).
    }
    \begin{threeparttable}
    \begin{NiceTabular*}{\textwidth}{@{\extracolsep{\fill}} c ccc c ccccc cccc @{}}
        \toprule[1.25pt]
        \multirow{2}{*}{\textbf{\rotatebox[origin=c]{90}{Model}}}
        & \multicolumn{3}{c}{\textsc{\textbf{Stage} 1}}
        & \multicolumn{1}{c}{\textsc{\textbf{Stage} 2}} 
        & \multicolumn{5}{c}{\textbf{NLG Metrics}} 
        & \multicolumn{4}{c}{\textbf{CE Metrics}} \\
        
        \cmidrule(lr){2-4}
        \cmidrule(lr){5-5}
        \cmidrule(lr){6-10}
        \cmidrule(lr){11-14}
        
        & $\mathrm{\textbf{PrAM}}$
        & $\boldsymbol{V_{ch}\!-\!T_{ch}}$
        & $\boldsymbol{L_{ch}}$
        & $\boldsymbol{\mathcal{G}_\mathrm{med}}$
        & \textbf{B1} & \textbf{B4} & \textbf{MTR} & \textbf{RG-L} & $\nabla_\mathrm{N}$
        & \textbf{Pre} & \textbf{Rec} & \textbf{F1} & $\nabla_\mathrm{C}$ \\
        \midrule
        \midrule
        Base$^\text{x}$
        & $\textcolor{red!50}{\times}$ & $\textcolor{red!50}{\times}$ & $\textcolor{red!50}{\times}$ & $\textcolor{red!50}{\times}$ 
        & 0.381$\pm$0.004 & 0.185$\pm$0.005 & 0.157$\pm$0.003 & 0.295$\pm$0.004 
        & -- & 0.455 & 0.471 & 0.463 & -- \\
        
        (a) 
        & $\textcolor{red!50}{\times}$ & \textcolor{forestgreen}{\checkmark} & $\textcolor{red!50}{\times}$ & $\textcolor{red!50}{\times}$ 
        & 0.399$\pm$0.003 & 0.197$\pm$0.004 & 0.160$\pm$0.002 & 0.301$\pm$0.005
        & 3.8\% & 0.472 & 0.493 & 0.482 & 4.2\% \\
        
        (b) 
        & \textcolor{forestgreen}{\checkmark} & $\textcolor{red!50}{\times}$ & $\textcolor{red!50}{\times}$ & $\textcolor{red!50}{\times}$ 
        & 0.404$\pm$0.005 & 0.208$\pm$0.006 & 0.163$\pm$0.003 & 0.302$\pm$0.004
        & 5.8\% & 0.485 & 0.501 & 0.493 & 6.5\% \\
        
        (c) 
        & \textcolor{forestgreen}{\checkmark} & \textcolor{forestgreen}{\checkmark} & $\textcolor{red!50}{\times}$ & $\textcolor{red!50}{\times}$ 
        & 0.422$\pm$0.004 & 0.233$\pm$0.005 & 0.169$\pm$0.004 & 0.311$\pm$0.006 
        & 11.5\% & 0.498 & 0.519 & 0.508 & 9.8\% \\
            
        (d) 
        & $\textcolor{red!50}{\times}$ & \textcolor{forestgreen}{\checkmark} & \textcolor{forestgreen}{\checkmark} & $\textcolor{red!50}{\times}$ 
        & 0.442$\pm$0.006 & 0.224$\pm$0.004 & 0.169$\pm$0.003 & 0.309$\pm$0.005 
        & 12.4\% & 0.511 & 0.541 & 0.526 & 13.6\% \\

        (e) 
        & \textcolor{forestgreen}{\checkmark} & \textcolor{forestgreen}{\checkmark} & \textcolor{forestgreen}{\checkmark} & $\textcolor{red!50}{\times}$ 
        & 0.465$\pm$0.005 & 0.247$\pm$0.006 & 0.175$\pm$0.004 & 0.314$\pm$0.004 
        & 18.0\% & 0.542 & 0.565 & 0.553 & 19.5\% \\

        (f) 
        & $\textcolor{red!50}{\times}$ & \textcolor{forestgreen}{\checkmark} & \textcolor{forestgreen}{\checkmark} & \textcolor{forestgreen}{\checkmark} 
        & 0.481$\pm$0.006 & 0.262$\pm$0.001 & 0.183$\pm$0.005 & 0.325$\pm$0.004 
        & 22.8\% & 0.563 & 0.592 & 0.577 & 24.6\% \\
        
        \cdashline{1-14}[0.5pt/2pt]\noalign{\vskip 0.5ex}
        \rowcolor{black!7}
        (g)$^\text{y}$
        & \textcolor{forestgreen}{\checkmark} & \textcolor{forestgreen}{\checkmark} & \textcolor{forestgreen}{\checkmark} & \textcolor{forestgreen}{\checkmark} 
        & 0.502$\pm$0.003$^*$ & 0.283$\pm$0.002$^*$ & 0.193$\pm$0.003$^*$ & 0.336$\pm$0.006$^*$ 
        & 29.0\% & 0.601 & 0.621 & 0.610 & 31.9\% \\
        
        \bottomrule[1.25pt]
    \end{NiceTabular*}
    \begin{tablenotes}
    \scriptsize
    \centering
    \item PrAM: Prior-Guided Attention Module; $V_{ch}$: Visual Channel; $T_{ch}$: Textual Channel; $L_{ch}$: Predicted diagnostic label; $\mathcal{G}_\mathrm{med}$: Graph-guided retrieval.
    \item $\nabla_\mathrm{N}$ and $\nabla_\mathrm{C}$: Average improvement over the base configuration for NLG and CE metrics; $\text{x}$: CliQ-RRG (Base Model); $\text{y}$: CliQ-RRG (Graph-guided)
    \end{tablenotes}
    \end{threeparttable}
    \label{tab:ablation}
\end{table*}

\section{Result and Discussion}
\noindent\textbf{Quantitative Analysis:}
\label{sec:qualitative}
We compare CliQ-RRG against seven method categories (Appendix \ref{sec:baselines}) on MIMIC-CXR and IU X-Ray. Reproduced baselines ($\star$) follow their original settings and evaluation protocols, without adding prior scans or external knowledge. For fair evaluation, we consider two settings: (i) CliQ-RRG (Graph-guided), where narrative reports are evaluated using standard NLG and CE metrics; and (ii) QA-Pair ($R_{qa}$). Since QA outputs differ from narrative reports, conventional NLG metrics are not directly applicable. We therefore evaluate QA outputs using semantic (BERTScore) and clinical (CheXbert-F1, RadGraph-F1) metrics, following prior QA-style evaluation practice~\cite{Manes2024kqa,Kim2024medexqa}. However, because QA-style text explicitly states disease names and their assertions, these metrics may not be directly comparable to scores on narrative reports. We therefore use \hyperlink{expert}{Expert Evaluation} as the primary evidence of clinical validity for the generated QA pairs.

\smallskip
\noindent\textit{Comparison with SOTA:}
As shown in Table~\ref{tab:sota}, CliQ-RRG achieves SOTA performance across most metrics on MIMIC-CXR and IU X-Ray. Compared with contrastive methods, our framework shows stronger lexical and clinical alignment with GT reports. Among knowledge-based methods, RADAR~\cite{Hou2025radar} remains competitive, while our graph-guided retrieval achieves superior overall performance. Among knowledge-based methods, REVTAF~\cite{Zhou2025revtaf} achieves the best MTR and second-best RG-L on MIMIC-CXR, while DAD-RG~\cite{Wu26dadsrg} performs strongly on IU X-Ray. CliQ-RRG outperforms multi-view methods, including MPO~\cite{Xiao2025mpo} and QA-style baseline CoD~\cite{Jin2025codvqa}, demonstrating the effectiveness of PrAM, channel alignment, and graph-guided retrieval. Notably, CliQ-RRG also outperforms the recent RA-RRG~\cite{Park2026rarrg} and MLLM-based MAIRA-2 \cite{bannur2024maira} and MedGemma \cite{sellergren2025medgemma} across the reported metrics.

\begin{figure*}[!ht]
\centering
    \includegraphics[width=0.9\textwidth]{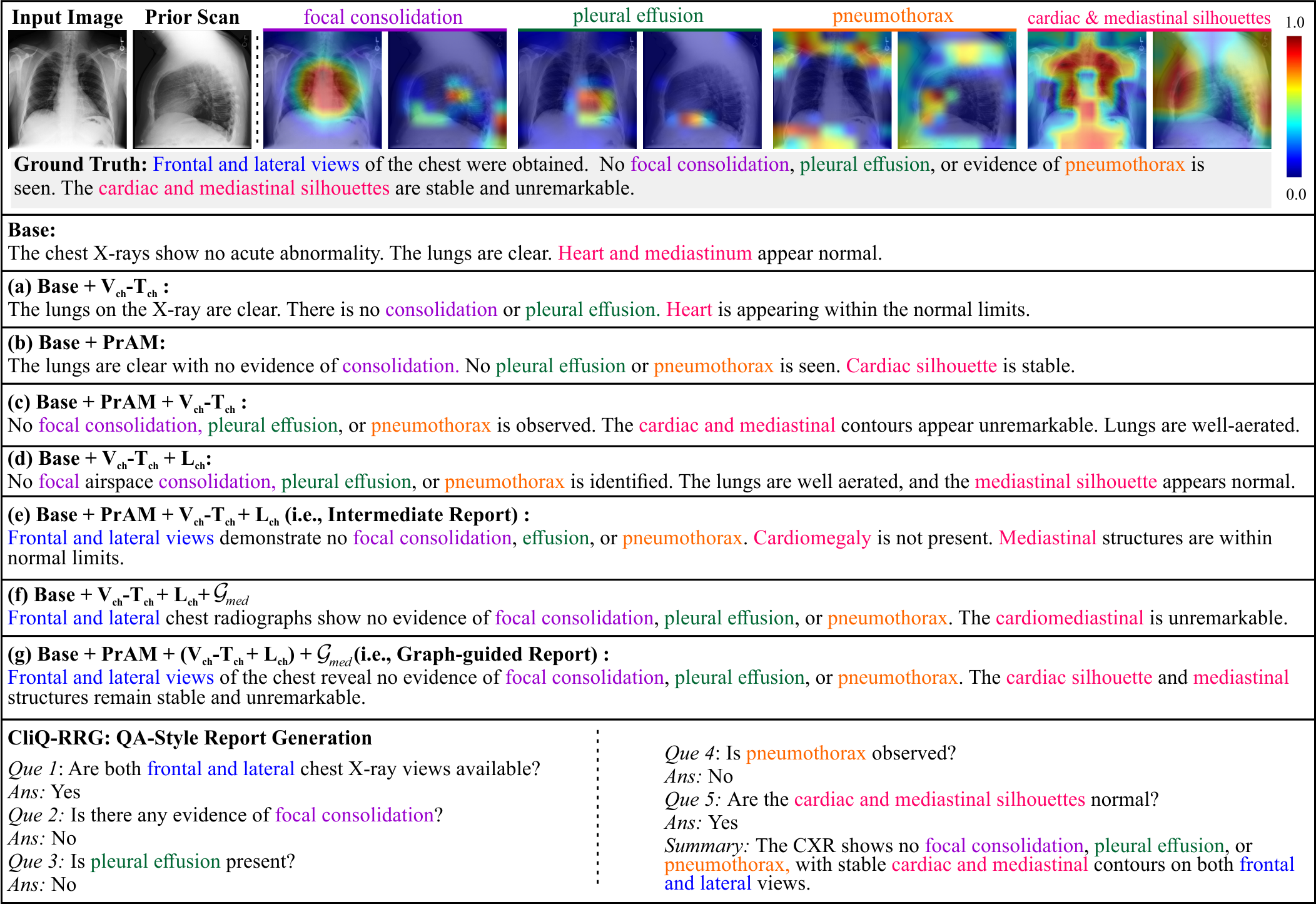} 
    \caption{Qualitative analysis of CliQ-RRG on MIMIC-CXR. Report comparison across models given input and prior scans, with ground truth (GT). Shared medical terms use consistent color coding, and Grad-CAM highlights anatomically and pathologically relevant regions.}  
\label{fig:qualitative}
\end{figure*}

\smallskip
\noindent\textit{CE Evaluation:}
Table~\ref{tab:ce} reports CE metrics on MIMIC-CXR, where CliQ-RRG achieves superior performance and outperforms REVTAF \cite{Zhou2025revtaf} in Rec and F1. We further evaluate the disease classifier, confirming its effectiveness in guiding multimodal alignment.

\smallskip
\noindent\textit{QA-pair Evaluation:}
Table~\ref{tab:qaeval} reports QA-pair evaluation on MIMIC-CXR and IU X-Ray. CliQ-RRG achieves strong BERTScore, CheXbert-F1, and RadGraph-F1 performance, indicating semantic fidelity and clinical consistency in QA pairs.
\begin{table}[!ht]
\scriptsize
\centering
\caption{QA-pair evaluation using semantic and clinical metrics.}
\begin{NiceTabular*}{\columnwidth}{@{\extracolsep{\fill}} lccc}
\toprule[1.25pt]
\textbf{Dataset} & \textbf{BERTScore} & \textbf{CheXbert-F1} & \textbf{RadGraph-F1} \\
\midrule\midrule
MIMIC-CXR & 0.921 & 0.639 & 0.325 \\
IU X-Ray  & 0.903 & 0.445 & 0.267 \\
\bottomrule[1.25pt]
\end{NiceTabular*}
\vspace{-10pt}
\label{tab:qaeval}
\end{table}

\smallskip
\noindent\textbf{Ablation Study:}
We analyze each component of CliQ-RRG on MIMIC-CXR (Table~\ref{tab:ablation}) and IU X-Ray (Appendix~\ref{sec:abl-iu}). The \textbf{base} model uses a visual encoder and Transformer decoder without alignment or graph-guided retrieval. Integrating visual ($V_{ch}$) and textual ($T_{ch}$) channels \textbf{(model a)} yields marginal gains (B4: $0.197$, F1: $0.482$), showing the limitation of CLIP-style alignment. Adding PrAM \textbf{(model b)} improves clinical performance (F1: $0.493$). Combining both \textbf{(model c)} further improves F1 to $0.508$. Replacing CLIP-style alignment with Di-VTC \textbf{(model d)}, which aligns $V_{ch}$--$T_{ch}$--$L_{ch}$, yields stronger gains (B4: $0.224$), showing the benefit of label-aware alignment. Integrating PrAM with Di-VTC \textbf{(model e)} further improves NLG and CE metrics. Adding $\mathcal{G}_{med}$ \textbf{(model f)} improves semantic completeness (F1: $0.577$). The full graph-guided framework \textbf{(model g)} achieves the best performance (NLG: $+29.0\%$, CE: $+31.9\%$), validating that $\mathcal{G}_{med}$ complements $V_{ch}{-}T_{ch}{-}L_{ch}$ alignment. Results are reported as mean $\pm$ std over five runs, and model (g) shows statistically significant gains ($p\le0.05$). We further analyze LLM-based QA restructuring and $\mathcal{G}_{med}$ in Appendix~\ref{sec:abl-llm} and~\ref{sec:abl-graphrag}.

\smallskip
\noindent\textbf{Qualitative Analysis;}
Fig.~\ref{fig:qualitative} illustrates how each component of CliQ-RRG progressively enhances report quality on a MIMIC-CXR sample. The base model produces generic statements such as \textit{“no acute abnormality”}, lacking clinical detail. Adding the \textit{PrAM} introduces precise findings (\textit{“no evidence of consolidation”}), while the \textit{Disease-aware Visual–Textual Contrast} identifies key observations like \textit{“focal consolidation, pleural effusion, or pneumothorax”}. Incorporating predicted disease embeddings improves diagnostic grounding, capturing terms such as \textit{“cardiomegaly”}. Graph-guided retrieval further refines report quality through clinically grounded evidence, producing findings such as \textit{“stable and unremarkable”}, aligning more closely with the ground truth and enabling structured QA-style outputs. GradCAM~\cite{Selvaraju2017gradcam} visualizations confirm that CliQ-RRG focuses on relevant anatomical regions, validating anatomically grounded reasoning.

\begin{figure}[!ht]
    \centering
    \includegraphics[width=\columnwidth]{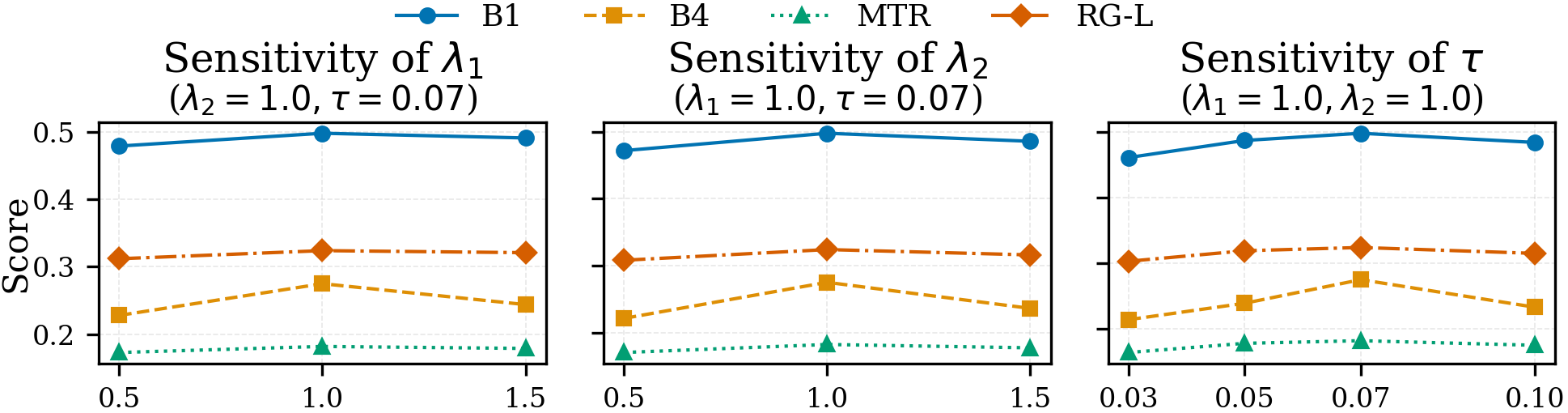}
    \caption{Parameter sensitivity analysis on MIMIC-CXR showing the effect of generation loss weight ($\lambda_1$), classification loss weight ($\lambda_2$), and temperature ($\tau$).}
    \label{fig:sensitivity}
\end{figure}

\smallskip
\noindent\textbf{Parameter Sensitivity Analysis:}
We analyze the sensitivity of key hyperparameters $\lambda_1$, $\lambda_2$, and $\tau$ on MIMIC-CXR. As shown in Fig.~\ref{fig:sensitivity}, we vary $\lambda_1, \lambda_2 {\in} {0.5, 1.0, 1.5}$ and $\tau {\in} {0.03, 0.05, 0.07, 0.10}$ while keeping other parameters fixed. The method achieves optimal performance across all metrics when $\lambda_1{=}\lambda_2{=}1.0$, indicating that equal weighting balances generation and classification objectives. Performance peaks at $\tau=0.07$, suggesting an appropriate temperature for contrastive alignment.

\smallskip
\hypertarget{expert}{}
\noindent\textbf{Expert Evaluation:}
We conduct a two-tier clinical validation to assess the reliability of the generated QA pairs. (i) We randomly sample 1,000 QA pairs generated by CliQ-RRG and evaluate using a standard Likert-scale protocol. Three medical professionals (2 doctors and 1 medical student) independently rate each QA pair against the ground-truth report on a 5-point scale across five criteria: overall quality, consistency, clinical relevance, specificity, and fluency, following the rubric of \citet{Hashemi2024llmrubric} (Table~\ref{tab:human_eval} (i)). We measure annotation reliability using Fleiss’ kappa $(\kappa_F)$ and pairwise agreement (PA). (ii) To assess clinical accuracy and hallucination, we perform an independent clinical validation with a senior radiologist with over 20 years of experience. We select 500 QA pairs using stratified random sampling across the 13 CheXbert disease categories and difficulty levels, and present them in randomized order without revealing their source. Each pair is compared with the corresponding GT report and categorized as fully acceptable, clinically acceptable, or hallucinated. As shown in Table~\ref{tab:human_eval} (ii), 467 pairs are fully acceptable, 31 are clinically acceptable, and 2 are hallucinated, yielding a 99.6\% clinical reliability rate and a 0.4\% hallucination rate. The detailed sampling protocol, difficulty distribution, error taxonomy, and inter-rater agreement are provided in Appendix~\ref{sec:expert_appendix}.

\begin{table}[!ht]
    \scriptsize
    \setlength{\tabcolsep}{3pt}
    \centering
    \caption{Human evaluation of generated QA pairs: (i) 5-point Likert ratings ($R_0$–$R_4$) with inter-annotator agreement; (ii) independent clinical assessment of QA integrity and hallucinations against ground truth.}
    \begin{NiceTabular*}{\columnwidth}{@{\extracolsep{\fill}} l@{}l ccccc | cc @{}}
    \toprule[1.25pt]
    \multirow{3}{*}{\textbf{(i)}} 
     & \textbf{Dataset} & $\mathbf{R_0}$ & $\mathbf{R_1}$ & $\mathbf{R_2}$ & $\mathbf{R_3}$ & $\mathbf{R_4}$ & \textbf{PA} & $\boldsymbol{\kappa_F}$ \\
    
    \cmidrule(lr){2-9}\morecmidrules\cmidrule(lr){2-9}
    
    & MIMIC-CXR & 4.07 & 4.03 & 4.32 & 4.17 & 4.29 & 0.83 & 0.77 \\    
    & IU X-Ray  & 4.15 & 4.07 & 4.51 & 3.97 & 4.21 & 0.86 & 0.79 \\    
    
    \bottomrule[1.25pt]
    \end{NiceTabular*}

    \vspace{2pt}

    \begin{NiceTabular*}{\columnwidth}{@{\extracolsep{\fill}} l@{}l c c @{}}
    \multirow{3}{*}{\textbf{(ii)}} 
     & \textbf{Outcome} & \textbf{Count} & \textbf{\%} \\
    
    \cmidrule(lr){2-4}\morecmidrules\cmidrule(lr){2-4}
    
    & Fully acceptable & 467 & 93.4\% \\    
    & Clinically acceptable (minor linguistic deviations) & 31 & 6.2\% \\    
    & Hallucinated (contradictory/unsupported findings) & 2 & 0.4\% \\
    
    \cdashline{2-4}[0.5pt/2pt]\noalign{\vskip 0.5ex}
    
    & \textbf{Total Reliable (Fully + Clinically) Pairs} & \textbf{498} & \textbf{99.6\%} \\
    
    \bottomrule[1.25pt]
    \end{NiceTabular*}

    \label{tab:human_eval}
\end{table}

\noindent\textbf{Generalizability Analysis:}
To assess generalization under distribution shift, we evaluate CliQ-RRG zero-shot on the held-out ReXGradient-160K~\cite{Zhang2025rexgradient}, a large-scale multi-institutional CXR dataset spanning 3 health systems and 79 medical sites. Neither training nor fine-tuning is performed on this dataset. Table~\ref{tab:rexgradient_nlg} reports Stage~1 results, while Table~\ref{tab:rexgradient_qa} reports the final QA-pair performance. Models trained on MIMIC-CXR consistently outperform those trained on IU X-Ray, reflecting the benefit of larger and more diverse training data. Although surface-level metrics show a moderate decrease under distribution shift, semantic and clinical metrics remain relatively stable, indicating that CliQ-RRG preserves clinical information across institutions and devices. This evaluation supports cross-hospital and cross-device generalization, while extension to other anatomical regions and modalities remains future work.

\begin{table}[!ht]
\scriptsize
\setlength{\tabcolsep}{2pt}
\centering
\caption{Zero-shot generalization of Stage~1 on ReXGradient-160K.}
\begin{NiceTabular*}{\columnwidth}{@{\extracolsep{\fill}} lccccccc}
\toprule[1.25pt]
\textbf{Train} & \textbf{Test} & \textbf{B1} & \textbf{B4} & \textbf{MTR} & \textbf{RG-L} & \textbf{C} & \textbf{BS} \\
\midrule\midrule
MIMIC-CXR & ReXGradient & 0.441 & 0.221 & 0.161 & 0.288 & 0.312 & 0.861 \\
IU X-Ray & ReXGradient & 0.398 & 0.165 & 0.148 & 0.259 & 0.271 & 0.831 \\
\bottomrule[1.25pt]
\end{NiceTabular*}
\label{tab:rexgradient_nlg}
\end{table}

\begin{table}[!ht]
\scriptsize
\setlength{\tabcolsep}{2pt}
\centering
\caption{Zero-shot QA-pair evaluation on ReXGradient-160K.}
\begin{NiceTabular*}{\columnwidth}{@{\extracolsep{\fill}} lcccc}
\toprule[1.25pt]
\textbf{Train} & \textbf{Test} & \textbf{BERTScore} & \textbf{CheXbert-F1} & \textbf{RadGraph-F1} \\
\midrule\midrule
MIMIC-CXR & ReXGradient & 0.897 & 0.581 & 0.289 \\
IU X-Ray & ReXGradient & 0.869 & 0.502 & 0.243 \\
\bottomrule[1.25pt]
\end{NiceTabular*}
\label{tab:rexgradient_qa}
\end{table}

\section{Related Work}
Due to space constraints in the main text, we provide related work in Appendix \ref{sec:related}.

\section{Conclusion}
In this paper, we present CliQ-RRG for QA-style radiology report generation. CliQ-RRG introduces a disease-aware visual--textual contrastive scheme to align visual, textual, and predicted diagnostic labels, enhanced by prior-guided attention for integrating multi-view chest X-rays. Additionally, graph-guided retrieval incorporates clinically connected evidence to improve report quality and support structured QA generation. Qualitative analysis shows that CliQ-RRG effectively aligns visual, textual, and label modalities to generate clinically reliable QA-style radiology reports. Experiments on MIMIC-CXR and IU X-Ray benchmarks highlight the superiority of our framework over state-of-the-art methods.

\section{Limitations}
While CliQ-RRG demonstrates strong performance, several limitations suggest directions for future improvement. First, the current study focuses on 2D CXR modalities, which enable a controlled analysis of disease-aware alignment and QA-style reporting but do not cover modalities with richer spatial structure. Second, the large language model is used as a fixed restructuring component to convert graph-guided reports into structured yes or no QA pairs, rather than being jointly optimized for multimodal reasoning. Third, training is limited to MIMIC-CXR and IU X-Ray, which may constrain the learned representations to their specific acquisition protocols and population characteristics. Fourth, although PrAM incorporates historical scans, it does not explicitly encode temporal order or time intervals between examinations and therefore does not directly model longitudinal progression patterns. Finally, the framework relies on supervised learning with paired images, reports, and diagnostic labels, which may constrain scalability in data-scarce settings.

\bibliography{custom}

\appendix
\vspace{2pt}
\noindent\textbf{\large Appendix}
\vspace{5pt}

\noindent Related Work \dotfill Appendix~\ref{sec:related} \\
Compared Baselines \dotfill Appendix~\ref{sec:baselines} \\
Knowledge Base Construction \dotfill Appendix~\ref{sec:knowledgebase} \\
Graph-guided Retrieval Details \dotfill Appendix~\ref{sec:graphretrieval}
Implementation Details \dotfill Appendix~\ref{sec:implementation} \\
Ablation Results (IU dataset) \dotfill Appendix~\ref{sec:abl-iu} \\
LLM Choice on QA \dotfill Appendix~\ref{sec:abl-llm} \\
Impact of Graph-retrival \dotfill Appendix~\ref{sec:abl-graphrag}\\
Per-State Disease Classification \dotfill Appendix~\ref{sec:perstate} \\
Noise-injection Sensitivity \dotfill Appendix~\ref{sec:noise}\\
Expanded Expert Evaluation \dotfill Appendix~\ref{sec:expert_appendix} \\
Computational Efficiency \dotfill Appendix~\ref{sec:computation} \\
Practical Implication \dotfill Appendix~\ref{sec:practical} \\
Future Work \dotfill Appendix~\ref{sec:future} \\
Reproducibility \dotfill Appendix \ref{sec:reproducibility}

\section{Related Work}
\label{sec:related}
\noindent\textbf{Encoder–Decoder and Multi-View RRG:}
Early RRG methods rely on encoder-decoder architectures to generate natural language descriptions from visual inputs \cite{Xu2015, Lu2017know, chen2020rr}. To better capture spatial and anatomical details, recent models incorporate multi-view X-rays and multi-slice features \cite{Chen2021cross, Bu2023refugen}. For longitudinal tracking, HC-LLM \cite{Liu2025hcllm} uses historical constraints to guide large language models in generating progression-aware reports.

\smallskip
\noindent\textbf{Contrastive Alignment:}
Multimodal contrastive learning is widely used to improve visual-textual alignment, ranging from global image-report matching to fine-grained region-word pairs \cite{Li2023dynamic, Shen2024man, Huangetal2025cmeaa}. Models like PromptMRG \cite{Jin2024promptmrg} use diagnostic labels as soft prompts to guide generation, while other methods integrate learned knowledge bases \cite{Yang2023radiology}. AdaMatch-Cyclic \cite{Chen2024adamatchcyclic} applies fine-grained cyclic alignment between image regions and text.

\smallskip
\noindent\textbf{Knowledge-Injected Generation:}
Recent works inject external clinical knowledge into the generation process via attention mechanisms and knowledge graphs \cite{Hou2025radar, Sun2025fact, Liu2024r2llm}. Notably, REVTAF \cite{Zhou2025revtaf} fuses modal alignment and knowledge injection using a learnable retrieval enhancer in hyperbolic space and optimal transport-based cross-attention. Similarly, KiUT \cite{Huang2023kiut} integrates symptom graphs into a U-Transformer. Other approaches, like EKAGen \cite{Bu2024iekagen}, retrieve instance-level expert knowledge and highlight key pathological regions, while DKP \cite{bu2024dkp} generates dynamic knowledge prompts from anomaly-driven features.

\smallskip
\noindent\textbf{QA-Style Generation:}
Motivated by limitations of narrative reporting \cite{Jin2024promptmrg, Luo2024textual, Wang2025llmrg4, kell2024question, pattnayak2025clinical, schwartz2011improving}, limited clinical grounding \cite{Yan2023see, Bu2024iekagen, Gu2024complex}, underuse of diagnostic labels \cite{Li2023dynamic, Shen2024man, Xiao2025mpo}, and limited temporal fusion \cite{Hou2023organ, Gu2025orid}, recent work explores structured and explainable outputs. For example, CoD \cite{Jin2025codvqa} prompts LLMs using findings extracted from diagnostic QA pairs. In medical VQA, MENDER \cite{Lin2025mendervqa} leverages cross-modal knowledge diffusion to achieve accurate answering. VED \cite{marhuenda2025unveilingvqa} uses a vision encoder-decoder to detect and explain radiological changes across longitudinal X-rays.

\smallskip
\noindent\textbf{Summary:}
Despite progress in alignment, knowledge injection, and QA formatting, prior works largely overlook longitudinal priors for structured reporting. CliQ-RRG addresses this by unifying disease-aware contrastive alignment with staged knowledge retrieval. Finally, it uses multi-view priors and employs an LLM to restructure standard narrative reports into precise, clinically meaningful QA-style outputs.

\section{Compared Baselines}
\label{sec:baselines}
We benchmark our proposed CliQ-RRG against state-of-the-art baselines, which we classify into seven categories.

\smallskip
\noindent\textbf{I: Contrastive-based Methods}
\begin{itemize}[noitemsep, leftmargin=*]
    \item\textit{R2GenRL} \cite{Qin2022r2genrl} optimizes the mapping between visual regions and textual words using reinforcement learning based on generation metrics.
    \item\textit{DCL} \cite{Li2023dynamic} constructs dynamic relation graphs and applies contrastive learning to align visual features with medical entities.
    \item\textit{MAN} \cite{Shen2024man} uses a shared memory mechanism to capture cross-modal correspondence and guide attention during decoding.
    \item\textit{CoFE} \cite{Li2024cofe} aligns representations via contrastive learning that maximize similarity for factual pairs while repelling negatives.
    \item\textit{SEI} \cite{Liu2024sei} aligns extracted anatomical entities and patient indications with corresponding visual regions.
    \item\textit{DART} \cite{Park2025dart} leverages diagnostic labels to ground text generation and incorporates a self-correcting mechanism to refine image-text consistency.
\end{itemize}

\smallskip
\noindent\textbf{II: Knowledge Injected-based Methods}
\begin{itemize}[noitemsep, leftmargin=*]
    \item\textit{KiUT} \cite{Huang2023kiut} integrates clinical knowledge into a U-Transformer architecture using a symptom graph and an adaptive distiller to guide word prediction.
    \item\textit{EKAGen} \cite{Bu2024iekagen} enhances generation by combining expert knowledge with discriminative attention mechanism to focus on pathological regions.    
    \item\textit{PromptMRG} \cite{Jin2024promptmrg} converts disease predictions into soft prompts and retrieves similar reports as in-context guidance.
    \item\textit{RADAR} \cite{Hou2025radar} injects external clinical knowledge and aligns it with visual features to enhance report reliability.     
     \item\textit{REVTAF} \cite{Zhou2025revtaf} utilizes semantic hierarchy in hyperbolic space to get reference reports to improve clinical findings.
\end{itemize}

\smallskip
\noindent{\textbf{III: LLM-based Methods}
\begin{itemize}[noitemsep, leftmargin=*]
    \item\textit{R2-LLM} \cite{Liu2024r2llm} frames report generation as instruction following using a frozen LLM guided by visual prompts.
    \item\textit{AdaMatch-Cyclic} \cite{Chen2024adamatchcyclic} employs a cyclic framework with adaptive patch–word alignment to guide both report generation and image synthesis.
    \item\textit{KARGEN} \cite{Li2024kargen} integrates a medical knowledge graph with a frozen LLM to generate disease-sensitive reports.
    \item\textit{LLM-RG4} \cite{Wang2025llmrg4} supports diverse input scenarios through adaptive token fusion and loss reweighting to reduce hallucinations.
    \item\textit{DAD-RG} \cite{Wu26dadsrg} utilizes a dual-attention framework that balances global radiological context with region-specific anatomical features to improve narrative consistency.
\end{itemize}

\smallskip
\noindent\textbf{IV: Multi View-based Methods}
\begin{itemize}[noitemsep, leftmargin=*]
    \item\textit{MPO} \cite{Xiao2025mpo}  adapts report generation to different user preferences by optimizing weighted objectives through reinforcement learning.
    \item\textit{HC-LLM} \cite{Liu2025hcllm} employs historical constraints to improve progression-aware report generation with large language models.
    \item\textit{ORID} \cite{Gu2025orid} filters irrelevant noise using an instruction-tuned LLaVA-Med model to generate organ-specific descriptions prioritized by graph-based analysis.
\end{itemize}

\smallskip
\noindent\textbf{V: QA Style-based Method}
\begin{itemize}[noitemsep, leftmargin=*]
    \item\textit{CoD} \cite{Jin2025codvqa} improves clinical accuracy by generating QA pairs through a diagnostic conversation framework to guide a large language model during RRG.
\end{itemize}

\smallskip
\noindent\textbf{VI: RAG-based Method}
\begin{itemize}[noitemsep, leftmargin=*]
    \item\textit{RA-RRG} \cite{Park2026rarrg} is a retrieval-augmented framework that extracts clinically essential key phrases and retrieves image-consistent phrases for hallucination-suppressed radiology report generation.
\end{itemize}

\smallskip
\noindent\textbf{VII: MLLM-based Methods}
\begin{itemize}[noitemsep, leftmargin=*]
    \item\textit{MAIRA-2} \cite{bannur2024maira} is a radiology-specific multimodal model that generates grounded and non-grounded chest X-ray reports using current and prior images, reports, and clinical context.
    \item\textit{MedGemma} \cite{sellergren2025medgemma} is a medical vision-language foundation model for interpreting medical images and text, with strong performance across medical VQA and chest X-ray report generation. 
\end{itemize}

\section{Medical Knowledge Base Construction}
\label{sec:knowledgebase}
We construct an external clinical repository
$\mathcal{K}_{ext}=\{K_m\}_{m=1}^{M}$
to support GraphRAG-based retrieval in \textsc{Stage} 2.

\noindent\textbf{Step 1: Source Collection:} We curate approximately 5,000 medical documents from PubMed abstracts related to thoracic imaging and chest radiology. Documents are filtered to retain clinically relevant evidence describing disease manifestations, imaging findings, and anatomical structures.

\noindent\textbf{Step 2: Clinical Fact Extraction:} Raw text is processed using BioClinicalBERT-based biomedical NER to identify radiology-relevant entities including diseases, symptoms, anatomical regions, and imaging observations. Sentences containing clinical entities are normalized into atomic knowledge units by removing redundant modifiers and standardizing terminology. Near-duplicate facts are filtered using cosine similarity thresholding.

\noindent\textbf{Step 3: Disease Label Association:} Each knowledge entry $K_m$ is associated with disease category
$y_m\in\mathcal{Y}$
to maintain consistency with predicted diagnostic labels
$L_{ch}$
obtained in \textsc{Stage} 1.

\noindent\textbf{Step 4: Embedding Initialization:} Knowledge units are embedded using BioWordVec to construct a searchable semantic space. The resulting embeddings support downstream graph grounding and retrieval.

\section{Graph-guided Retrieval Details}
\label{sec:graphretrieval}
\noindent\textbf{(i) Medical Graph Construction:}
GraphRAG is implemented using a medical knowledge graph $\mathcal{G}_{med}$ constructed from approximately 5,000 domain-specific medical summaries and clinical resources described in Appendix~\ref{sec:knowledgebase}. Graph entities represent diseases, anatomical structures, imaging findings, and clinical observations. 
Directed edges encode semantic medical relations $\mathcal{R}$, including \texttt{located\_in}, \texttt{associated\_with}, \texttt{manifestation\_of}, \texttt{suggestive\_of}, and \texttt{coexists\_with}.
Graph entities are pre-computed using BioWordVec embeddings to support downstream graph retrieval. Graph storage and traversal are implemented using Neo4j.

\noindent\textbf{(ii) Entity Grounding:}
During inference, intermediate report $R_I$ is encoded into report representation $\phi_{R_I}$, while predicted disease labels $L_{ch}$ provide diagnostic constraints. Entity grounding evaluates retrieval relevance using Eq.~\ref{eq:graphrank}. For diagnostic supervision, graph entities are mapped into the disease label space through a mapping operator $\mathcal{M}:\mathcal{V}\rightarrow \{1,2,\dots,|\mathcal{Y}|\}$, where $\mathcal{V}$ denotes graph entities and $\mathcal{Y}$ denotes diagnostic categories. Disease entities are mapped directly, while anatomical findings and imaging observations inherit labels through graph connectivity:
\begin{equation}
\mathcal{M}(v_i)=
\arg\max_{m\in\{1,\dots,|\mathcal{Y}|\}}
\mathrm{sim}
(\phi_{v_i},\phi_{y_m})
\end{equation}
where $\phi_{v_i}$ denotes graph entity embedding and $\phi_{y_m}$ denotes the embedding of disease category $y_m\in\mathcal{Y}$. Threshold $\tau\in[0,1]$ filters low-confidence disease predictions in $L_{ch}$, allowing only reliable diagnostic information to contribute to retrieval scoring. Top-$k$ ranked entities initialize retrieval seed set $\Lambda_s$, enabling graph retrieval using both report semantics and predicted disease labels.

\noindent\textbf{(iii) Subgraph Retrieval and Relation-path Discovery:}
Starting from $\Lambda_s$, bounded breadth-first traversal retrieves evidence subgraph $\mathcal{G}_{sub}$ while suppressing irrelevant evidence propagation. Traversal depth is restricted to two hops to preserve clinically relevant dependencies. Example relation paths include:
\begin{itemize}[leftmargin=*]
\item \texttt{Pleural Effusion $\rightarrow$ manifestation\_of $\rightarrow$ Costophrenic Angle Blunting}
\item \texttt{Opacity $\rightarrow$ located\_in $\rightarrow$ Left Lower Lobe}
\item \texttt{Cardiomegaly $\rightarrow$ associated\_with $\rightarrow$ Enlarged Cardiac Silhouette}
\end{itemize}

\noindent\textbf{(iv) Graph Context Serialization:}
Retrieved relation paths are ranked using retrieval relevance and diagnostic consistency. High-confidence paths are serialized into graph evidence context $C_G$ using structural delimiters while preserving relation ordering and graph connectivity. Serialized graph evidence is appended to intermediate report representation $R_I$ to obtain graph-enhanced context $R_{KI}$ for downstream QA-style generation.

\begin{figure*}[!ht]
    \centering
    \includegraphics[width=\textwidth]{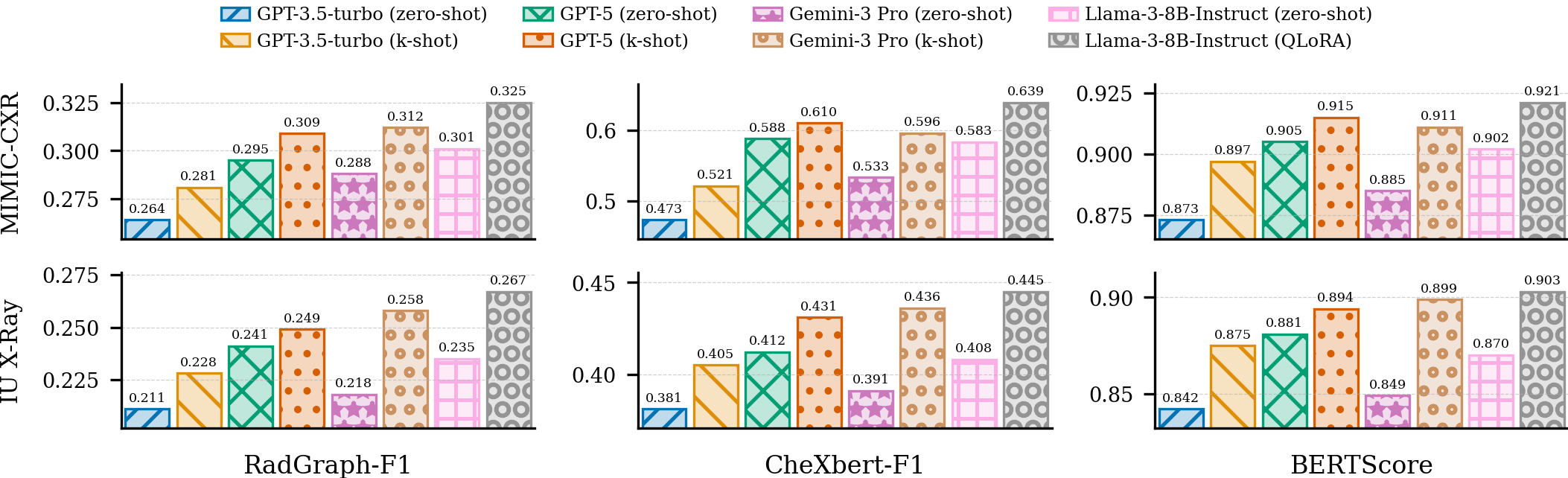}
    \caption{
    Comparison of LLMs for QA restructuring under zero-shot, few-shot, and QLoRA fine-tuned settings on MIMIC-CXR and IU X-Ray using RadGraph-F1, CheXbert-F1, and BERTScore.
    }
    \label{fig:llm}
\end{figure*}

\section{Implementation Details}
\label{sec:implementation}
Input CXRs are resized and cropped to $224\times224$ pixels. In \textsc{Stage~1}, visual and textual encoder outputs are projected to $d_e=768$ through a linear layer, while \textsc{Stage~2} utilizes a Transformer~\cite{Vaswani2017attention} decoder with $8$ attention heads and a hidden size of $256$. The generation length for $R_{KI}$ is capped at $100$ tokens. For GraphRAG retrieval, we construct a medical knowledge graph aligned with predicted disease labels (Appendix~\ref{sec:knowledgebase}). Graph entities are initialized using BioWordVec embeddings and stored within a Neo4j graph database. During inference, grounded entities initialize graph retrieval, where traversal is restricted to two hops for evidence subgraph extraction. We use $\mathrm{top\text{-}}k_t=10$ retrieval seeds for graph evidence retrieval.\footnote{We tested $\mathrm{top\text{-}}k_t\in[1,15]$ and selected $10$ based on validation performance.} QA restructuring is performed using Llama-3-8B-Instruct adapted through QLoRA fine-tuning with 4-bit NF4 quantization, LoRA rank $r=16$, scaling factor $\alpha=32$, and dropout $0.05$. Fine-tuning is performed using AdamW with a learning rate of $2\times10^{-4}$ for $3$ epochs. All experiments were supported by an Intel(R) Xeon(R) Silver 4316 CPU @ 2.30\,GHz with 512\,GB RAM, and inference was performed with $m=5$.

\section{Ablation Analysis on IU Dataset}
\label{sec:abl-iu}
We further analyze each component on IU X-Ray (Table~\ref{tab:abl-iu}). The \textbf{base} encoder--decoder achieves a B1 score of $0.435$. Integrating visual ($V_{ch}$) and textual ($T_{ch}$) channels in \textbf{model (a)} improves average performance by $5.4\%$, showing the limitation of standard visual--textual alignment. Adding PrAM in \textbf{model (b)} yields an $8.2\%$ gain, highlighting the benefit of multi-view context. Combining both in \textbf{model (c)} further improves performance. Replacing standard alignment with Di-VTC in \textbf{model (d)}, which aligns $V_{ch}$--$T_{ch}$--$L_{ch}$, improves B1 to $0.498$, demonstrating stronger label-aware grounding. Integrating PrAM with Di-VTC in \textbf{model (e)} further improves NLG performance. Adding $\mathcal{G}_{med}$ in \textbf{model (f)} enhances clinical completeness. The full graph-guided framework \textbf{(model g)} achieves the best performance ($+27.1\%$), confirming the effectiveness of graph-guided retrieval. These observations remain consistent with MIMIC-CXR and demonstrate the robustness of CliQ-RRG.

\begin{table}[!ht]
    \tiny
    \setlength{\tabcolsep}{2pt}
    \centering
    \caption{Ablation analysis on IU X-Ray showing incremental gains from integrating PrAM, Di-VTC (across $V_{ch}$--$T_{ch}$ and $L_{ch}$), and graph-guided retrieval ($\mathcal{G}_{med}$) into the base model.}
    \begin{threeparttable}
    \begin{NiceTabular*}{\columnwidth}{@{\extracolsep{\fill}} c @{} c@{}c@{}c c cccc c @{}}
        \toprule[1.25pt]

        \multirow{2}{*}{\textbf{\rotatebox[origin=c]{90}{Model}}}
        & \multicolumn{3}{c}{\textsc{\textbf{Stage} 1}}
        & \multicolumn{1}{c}{\textsc{\textbf{Stage} 2}}
        & \multicolumn{5}{c}{\textbf{NLG Metrics}} \\

        \cmidrule(lr){2-4}
        \cmidrule(lr){5-5}
        \cmidrule(lr){6-10}

        & $\mathrm{\textbf{PrAM}}$
        & $\boldsymbol{V_{ch}\!-\!T_{ch}}$
        & $\boldsymbol{L_{ch}}$
        & $\boldsymbol{\mathcal{G}_{med}}$
        & \textbf{B1}
        & \textbf{B4}
        & \textbf{MTR}
        & \textbf{RG-L}
        & $\nabla_\mathrm{N}$ \\

        \midrule
        \midrule

        Base
        & $\textcolor{red!50}{\times}$
        & $\textcolor{red!50}{\times}$
        & $\textcolor{red!50}{\times}$
        & $\textcolor{red!50}{\times}$
        & 0.435
        & 0.144
        & 0.174
        & 0.387
        & -- \\

        (a)
        & $\textcolor{red!50}{\times}$
        & \textcolor{forestgreen}{\checkmark}
        & $\textcolor{red!50}{\times}$
        & $\textcolor{red!50}{\times}$
        & 0.452
        & 0.168
        & 0.184
        & 0.398
        & 5.4\% \\

        (b)
        & \textcolor{forestgreen}{\checkmark}
        & $\textcolor{red!50}{\times}$
        & $\textcolor{red!50}{\times}$
        & $\textcolor{red!50}{\times}$
        & 0.468
        & 0.165
        & 0.196
        & 0.404
        & 8.2\% \\

        (c)
        & \textcolor{forestgreen}{\checkmark}
        & \textcolor{forestgreen}{\checkmark}
        & $\textcolor{red!50}{\times}$
        & $\textcolor{red!50}{\times}$
        & 0.489
        & 0.194
        & 0.205
        & 0.418
        & 14.6\% \\

        (d)
        & $\textcolor{red!50}{\times}$
        & \textcolor{forestgreen}{\checkmark}
        & \textcolor{forestgreen}{\checkmark}
        & $\textcolor{red!50}{\times}$
        & 0.498
        & 0.188
        & 0.201
        & 0.411
        & 13.9\% \\

        (e)
        & \textcolor{forestgreen}{\checkmark}
        & \textcolor{forestgreen}{\checkmark}
        & \textcolor{forestgreen}{\checkmark}
        & $\textcolor{red!50}{\times}$
        & 0.531
        & 0.207
        & 0.213
        & 0.425
        & 20.7\% \\

        (f)
        & $\textcolor{red!50}{\times}$
        & \textcolor{forestgreen}{\checkmark}
        & \textcolor{forestgreen}{\checkmark}
        & \textcolor{forestgreen}{\checkmark}
        & 0.551
        & 0.217
        & 0.221
        & 0.432
        & 24.6\% \\

        \cdashline{1-10}[0.5pt/2pt]\noalign{\vskip 0.5ex}

        \rowcolor{black!7}
        (g)
        & \textcolor{forestgreen}{\checkmark}
        & \textcolor{forestgreen}{\checkmark}
        & \textcolor{forestgreen}{\checkmark}
        & \textcolor{forestgreen}{\checkmark}
        & \textbf{0.565}
        & \textbf{0.221}
        & 0.225
        & 0.438
        & 27.1\% \\

        \bottomrule[1.25pt]

    \end{NiceTabular*}

    \begin{tablenotes}
    \scriptsize
    \centering

    \item PrAM: Prior-Guided Attention Module;
    $V_{ch}$: Visual Channel;
    $T_{ch}$: Textual Channel;
    $L_{ch}$: Predicted diagnostic label;
    $\mathcal{G}_{med}$: Graph-guided retrieval.

    \item $\nabla_\mathrm{N}$:
    Average improvement across NLG metrics over the base model.

    \end{tablenotes}

    \end{threeparttable}
    \label{tab:abl-iu}

\end{table}

\section{Impact of LLMs on QA Restructuring}
\label{sec:abl-llm}
To assess QA restructuring, Fig.~\ref{fig:llm} compares multiple language models across prompting and fine-tuning settings using RadGraph-F1, CheXbert-F1, and BERTScore. Few-shot prompting ($k=5$) improves over zero-shot settings across different models. Fine-tuned Llama-3-8B-Instruct with QLoRA achieves the best overall performance, outperforming larger closed-source models such as GPT-5 and Gemini-3 Pro. These findings suggest that lightweight task adaptation is more effective than prompt design alone for preserving clinical entities and semantic alignment.

\section{Impact of Graph-retrieval}
\label{sec:abl-graphrag}
We analyze the contribution of graph-guided retrieval (see Table~\ref{tab:graph-impact}) through retrieval seed initialization ($\Lambda_s$), relation-path retrieval ($p_r$), graph serialization ($C_G$), and traversal depth (Hop). Removing $\Lambda_s$ causes the largest performance drop, reducing CheXbert-F1 from $0.610$ to $0.591$, highlighting the importance of grounded retrieval initialization. Removing $C_G$ limits sequential coherence also degrades performance, resulting in a severe drop in RG-L to 0.299 indicating that structured graph evidence improves report generation quality. Traversal analysis shows that 2-hop retrieval achieves the best performance, while 1-hop limits evidence coverage and 3-hop introduces irrelevant anatomical associations that degrade generation quality across all lexical and clinical metrics.

\begin{table}[!ht]
\scriptsize
\setlength{\tabcolsep}{2pt}
\centering

\caption{Impact of graph-guided retrieval on MIMIC-CXR. Higher is better.}
\begin{threeparttable}
\begin{NiceTabular*}{\columnwidth}
{@{\extracolsep{\fill}} l @{} c@{}c@{}c c ccc @{}}
\toprule[1.25pt]
\textbf{Setting}
&
$\Lambda_s$ & $p_r$ & $C_G$
& \textbf{Hop} & \textbf{B4} & \textbf{RG-L} & \textbf{F1} \\

\midrule\midrule

w/o $\Lambda_s$
& $\textcolor{red!50}{\times}$
& \textcolor{forestgreen}{\checkmark}
& \textcolor{forestgreen}{\checkmark}
& 2 & 0.242 & 0.311 & 0.548 \\

w/o $p_r$
& \textcolor{forestgreen}{\checkmark}
& $\textcolor{red!50}{\times}$
& \textcolor{forestgreen}{\checkmark}
& 2 & 0.259 & 0.318 & 0.574 \\

w/o $C_G$
& \textcolor{forestgreen}{\checkmark}
& \textcolor{forestgreen}{\checkmark}
& $\textcolor{red!50}{\times}$
& 2 & 0.248 & 0.299 & 0.561 \\

1-hop
& \textcolor{forestgreen}{\checkmark}
& \textcolor{forestgreen}{\checkmark}
& \textcolor{forestgreen}{\checkmark}
& 1 & 0.270 & 0.324 & 0.589 \\

3-hop
& \textcolor{forestgreen}{\checkmark}
& \textcolor{forestgreen}{\checkmark}
& \textcolor{forestgreen}{\checkmark}
& 3 & 0.276 & 0.329 & 0.595 \\

\cdashline{1-8}[0.5pt/2pt]\noalign{\vskip 0.5ex}
\rowcolor{black!7}

2-hop (Full)
& \textcolor{forestgreen}{\checkmark}
& \textcolor{forestgreen}{\checkmark}
& \textcolor{forestgreen}{\checkmark}
& 2 & \textbf{0.283} & \textbf{0.336} & \textbf{0.610} \\

\bottomrule[1.25pt]
\end{NiceTabular*}
\begin{tablenotes}
\scriptsize
\centering
\item
$\Lambda_s$: retrieval seed initialization;
$p_r$: relation-path retrieval;\\
$C_G$: graph context serialization
\end{tablenotes}
\end{threeparttable}
\label{tab:graph-impact}
\end{table}

\section{Per-State Disease Classification}
\label{sec:perstate}
To verify that the disease classifier captures the four-state labeling scheme used by CheXbert~\cite{Smit2020chexbert}, we report its per-state performance on MIMIC-CXR in Table~\ref{tab:perstate}. For each of the 13 disease categories, the classifier independently predicts one of four states: positive, negative, uncertain, or not mentioned. The reported values are averaged across the 13 disease categories. The classifier achieves a macro-average F1 of $0.505$, consistent with the overall disease-classification F1 reported in the main paper. Performance is higher for the positive and negative states, while the uncertain state is comparatively more challenging. This indicates that the classifier learns distinctions across the four CheXbert states rather than reducing the task to binary disease prediction.

\begin{table}[!ht]
\scriptsize
\centering
\caption{Per-state disease classification performance on MIMIC-CXR.}
\begin{NiceTabular*}{\columnwidth}{@{\extracolsep{\fill}} lcccc}
\toprule[1.25pt]
\textbf{State} & \textbf{Precision} & \textbf{Recall} & \textbf{F1} & \textbf{Support (\%)} \\
\midrule\midrule
Positive & 0.541 & 0.558 & 0.549 & 28.4 \\
Negative & 0.612 & 0.634 & 0.623 & 41.2 \\
Uncertain & 0.389 & 0.402 & 0.395 & 12.6 \\
Not Mentioned & 0.442 & 0.466 & 0.454 & 17.8 \\
\cdashline{1-5}[0.5pt/2pt]\noalign{\vskip 0.5ex}
\rowcolor{black!7}
\textbf{Macro-average} & \textbf{0.496} & \textbf{0.515} & \textbf{0.505} & \textbf{100.0} \\
\bottomrule[1.25pt]
\end{NiceTabular*}
\label{tab:perstate}
\end{table}

\section{Noise-injection Sensitivity}
\label{sec:noise}
To examine the sensitivity of Di-VTC to errors in the predicted disease labels $L_{ch}$, we conduct a controlled noise-injection study on MIMIC-CXR. During Stage~1 training, we synthetically corrupt the labels used to construct positive and negative pairs at rates of $0\%, 10\%, 20\%,$ and $30\%$, while keeping all other settings fixed. We then evaluate the resulting classifier, intermediate report generation, and final QA-pair quality. As shown in Table~\ref{tab:noise}, increasing label noise consistently degrades performance across all stages. At $20\%$ noise, B4 and QA-pair RadGraph-F1 decrease by $10.95\%$ and $10.46\%$, respectively, relative to the noise-free setting. At $30\%$ noise, the degradation becomes more pronounced, with B4 and RadGraph-F1 decreasing to $0.221$ and $0.256$, respectively. These results indicate that Di-VTC is moderately tolerant to label noise at low levels but becomes increasingly sensitive as classifier errors accumulate. Since the reported disease classifier achieves an F1 score of $0.505$, improving disease prediction remains an important direction for enhancing the reliability of disease-aware alignment and downstream generation.

\begin{table}[!ht]
\scriptsize
\centering
\caption{Sensitivity of Di-VTC and downstream generation to injected disease-label noise on MIMIC-CXR.}
\begin{NiceTabular*}{\columnwidth}{@{\extracolsep{\fill}} lcccc}
\toprule[1.25pt]
\textbf{Noise (\%)} & \textbf{F1} & \textbf{B4} & \textbf{RG-L} & \textbf{RadGraph-F1} \\
\midrule\midrule
30 \% & 0.402 & 0.221 & 0.278 & 0.256 \\

20 \% & 0.446 & 0.252 & 0.309 & 0.291 \\

10 \% & 0.481 & 0.271 & 0.327 & 0.314 \\

\cdashline{1-5}[0.5pt/2pt]\noalign{\vskip 0.5ex}
\rowcolor{black!7}

\textbf{0 \% (Ours)}  & \textbf{0.505} & \textbf{0.283} & \textbf{0.336} & \textbf{0.325} \\
\bottomrule[1.25pt]
\end{NiceTabular*}
\label{tab:noise}
\end{table}

\section{Expanded Expert Evaluation}
\label{sec:expert_appendix}
To further assess the clinical reliability of the generated QA pairs, we expanded the evaluation to 500 QA pairs from MIMIC-CXR. The samples were selected using stratified random sampling across the 13 CheXbert disease categories, with balanced representation of positive, negative, and uncertain findings, and were further grouped by difficulty (Table~\ref{tab:difficulty}). The QA pairs were presented in randomized order without revealing their source. Table~\ref{tab:error} summarizes the observed errors, with most involving minor linguistic deviations and only two cases (0.4\%) containing hallucinated findings. Inter-rater agreement is reported in Table~\ref{tab:human_eval}, with a pairwise agreement of 0.83 and Fleiss' $\kappa$ of 0.77, indicating substantial agreement and supporting the reliability of the clinical assessment.

\begin{table}[!ht]
\scriptsize
\centering
\caption{Difficulty distribution of the 500 QA pairs sampled for expert evaluation.}
\begin{NiceTabular*}{\columnwidth}{@{\extracolsep{\fill}} lccc}
\toprule[1.25pt]
\textbf{Difficulty} & \textbf{Definition} & \textbf{Count} & \textbf{(\%)} \\
\midrule\midrule
Easy & Single, unambiguous finding & 233 & 46.6 \\
Moderate & Single subtle/borderline finding & 187 & 37.4 \\
Hard & Multiple findings or rare pathology & 80 & 16.0 \\
\cdashline{1-4}[0.5pt/2pt]\noalign{\vskip 0.5ex}
\rowcolor{black!7}
\textbf{Total} & & \textbf{500} & \textbf{100.0} \\
\bottomrule[1.25pt]
\end{NiceTabular*}
\label{tab:difficulty}
\end{table}

\begin{table}[!ht]
\scriptsize
\centering
\caption{Error types identified among the non-fully-acceptable QA pairs.}
\begin{NiceTabular*}{\columnwidth}{@{\extracolsep{\fill}} lccc}
\toprule[1.25pt]
\textbf{Error Type} & \textbf{Count} & \textbf{Errors (\%)} & \textbf{Total (\%)} \\
\midrule\midrule
Minor linguistic deviation & 21 & 63.6 & 4.2 \\
Missing finding & 6 & 18.2 & 1.2 \\
Wrong laterality & 1 & 3.0 & 0.2 \\
Incorrect severity & 1 & 3.0 & 0.2 \\
Temporal inconsistency & 2 & 6.1 & 0.4 \\
Hallucinated finding & 2 & 6.1 & 0.4 \\
\cdashline{1-4}[0.5pt/2pt]\noalign{\vskip 0.5ex}
\rowcolor{black!7}
\textbf{Total} & \textbf{33} & \textbf{100.0} & \textbf{6.6} \\
\bottomrule[1.25pt]
\end{NiceTabular*}
\label{tab:error}
\end{table}

\section{Computational Efficiency}
\label{sec:computation}
We analyze the computational efficiency of CliQ-RRG on a single NVIDIA 64\,GB GPU, as summarized in Table~\ref{tab:efficiency}. The framework consists of a two-stage architecture with $\sim170$M trainable parameters, including \textsc{Stage}~1 multimodal alignment and \textsc{Stage}~2 graph-guided report generation. The average inference time is $180$--$300$\,ms per study, excluding the final QA restructuring step. Graph-guided retrieval introduces minimal overhead through lightweight entity grounding and graph traversal operations. QA restructuring is performed independently using QLoRA fine-tuned Llama-3-8B-Instruct, introducing an additional latency of $1.35$--$2.75$\,s per study.

\begin{table}[!ht]
\setlength{\tabcolsep}{2pt}
\scriptsize
\centering
\caption{Computational efficiency of CliQ-RRG.}
\begin{threeparttable}

\begin{NiceTabular*}{\columnwidth}{@{\extracolsep{\fill}} lcc}

\toprule[1.25pt]

\textbf{Method}
&
\textbf{Parameters}
&
\textbf{Inference Time}
\\

\midrule\midrule

CliQ-RRG (Graph-guided)
&
\multirow{2}{*}{$\sim170$M}
&
$180$--$300$\,ms / study$^{*}$
\\

CliQ-RRG (QA Pair)
&
&
$1.35$--$2.75$\,s / study$^{*}$
\\

\bottomrule[1.25pt]

\end{NiceTabular*}

\begin{tablenotes}
\scriptsize
\item
$*$ Reported for single-scan and prior-scan settings.
\end{tablenotes}

\end{threeparttable}

\label{tab:efficiency}

\end{table}

\section{Practical Implication}
\label{sec:practical}
CliQ-RRG offers several practical benefits for real-world clinical deployment and research use.
First, the QA-style report formulation converts free-text narratives into a structured format that is easier to read, review, and query, which supports faster clinical decision making.
Second, the disease-aware design ties each reported finding to explicit diagnostic evidence, thereby improving report consistency and enabling clinicians to verify conclusions with greater confidence.
Third, the use of prior examinations supports longitudinal assessment and clearer tracking of disease changes across visits.
Fourth, the integration of clinical knowledge enriches reports with relevant medical context, improving coverage of main findings without increasing text density.
Finally, CliQ-RRG streamlines the radiology workflow by generating structured and reliable reports that clinicians can directly utilize.

\section{Future Work}
\label{sec:future}
Future work will focus on extending the framework along several methodological and practical dimensions. First, extending the framework toward Open-Set VQA generation is a promising direction for future work. Second, we will extend the framework to additional imaging modalities, including 3D data such as CT and MRI, to assess its applicability to volumetric and anatomically complex scenarios. We will also investigate explicit temporal modeling, such as time-interval embeddings, temporally ordered attention, or recurrent architectures, to better capture longitudinal progression across historical examinations. Third, we plan to explore vision--language models that directly perform multimodal question answering in an interactive manner. Fourth, we will investigate training on broader multi-center and cross-institutional datasets, along with prospective evaluation, to further assess robustness across diverse clinical environments. Fifth, we aim to investigate semi-supervised and weakly supervised learning strategies to reduce dependence on fully annotated data.

\section{Reproducibility}
\label{sec:reproducibility}
To support reproducibility, we release our code at: \url{https://github.com/chandravardhan-singh-raghaw/CliQ-RRG}

\end{document}